\documentclass[11pt]{article}

\usepackage[preprint]{acl}
\usepackage{times}
\usepackage{latexsym}
\usepackage{booktabs}

\usepackage[T1]{fontenc}
\usepackage[utf8]{inputenc}

\usepackage{microtype}

\usepackage{inconsolata}

\usepackage{graphicx}

\usepackage{amsmath}
\usepackage{amssymb}
\usepackage{algorithm}
\usepackage{algpseudocode}
\usepackage{hyperref}
\usepackage{tikz}
\usetikzlibrary{shapes,arrows,positioning,decorations.pathreplacing,fit}
\providecommand{\Description}[1]{\ignorespaces}

\title{Detecting Proxy Gaming in RL and LLM Alignment via Evaluator Stress Tests}

\author{
Ibne Farabi Shihab\thanks{Equal contribution.}\thanks{Corresponding author: \texttt{ishihab@iastate.edu}.}\textsuperscript{1}
\and
Sanjeda Akter\footnotemark[1]\textsuperscript{1}
\and
Anuj Sharma\textsuperscript{2}
\\[2pt]
\textsuperscript{1}Department of Computer Science, Iowa State University \\
\textsuperscript{2}Department of Civil, Construction \& Environmental Engineering, Iowa State University \\
\texttt{ishihab@iastate.edu}
}

\begin{document}
\maketitle

\begin{abstract}
Proxy optimization, where AI systems exploit evaluator weaknesses rather than improve intended objectives, threatens both reinforcement learning (reward hacking) and LLM alignment (evaluator gaming). We introduce the Evaluator Stress Test (EST), an invariance-based framework that detects proxy gaming by separating exploitable sensitivity (e.g., format, physics bugs) from content-driven improvements using controlled perturbations with semantic validity audits. We validate EST across both domains. In RL, across 15 environments and 5 algorithms (2,156 expert-annotated episodes), EST achieves 78.4\% precision and 81.7\% recall. In LLM alignment, across 4 tasks, 2 model scales, 2 training methods, and 2 judges (1,200 human-annotated instances), EST achieves 74.2\% precision and 78.6\% recall with early warning signals preceding quality decline. Cross-domain analysis reveals that proxy-true correlation tracking transfers directly between domains, while perturbation design requires domain adaptation. Closed-loop mitigation improves human win-rate by 8.3 points (LLM) and reduces hacking by 54.6\% (RL). We release benchmarks for both domains: 2,156 RL episodes and 1,200 LLM instances.
\end{abstract}

\section{Introduction}

Proxy optimization, where AI systems exploit measurable proxies rather than improve true objectives, is a fundamental challenge spanning reinforcement learning and LLM alignment~\cite{amodei2016concrete,russell2016research,kenton2021alignment}. In RL, agents optimize reward functions that imperfectly capture designer intent~\cite{sutton2018reinforcement,ng1999policy} across diverse applications including robotics~\cite{kober2013reinforcement}, autonomous driving~\cite{kiran2021deep}, financial systems~\cite{fischer2018reinforcement}, and game AI~\cite{vinyals2019grandmaster}, leading to reward hacking: behaviors like driving in circles to collect checkpoints or pausing indefinitely to avoid losing~\cite{krakovna2020specification,everitt2017reward}. This aligns with Goodhart's law~\cite{goodhart1984problems}, where metrics become unreliable when optimized. In LLM alignment, models optimize against learned reward models or LLM-as-judge evaluators~\cite{zheng2023judging}, leading to evaluator gaming: exploiting stylistic artifacts (verbosity, bullet points) that inflate scores without improving human preference~\cite{skalse2022defining}.

Despite surface differences, these failures share common structure: optimizers discover that proxy scores can increase without corresponding true objective improvement~\cite{amodei2016concrete,russell2016research}. This parallel suggests a unified detection approach. We test this hypothesis by introducing the Evaluator Stress Test (EST), an invariance-based framework that detects proxy gaming through controlled perturbations across both domains.

Legitimate improvements should be robust to perturbations that alter exploitable features while preserving task-relevant content. If an RL agent's high reward disappears when physics parameters change slightly, the agent was exploiting a simulation bug. If an LLM's high judge score disappears when output format changes, the model was exploiting format sensitivity. EST operationalizes this by measuring exploitable sensitivity, which captures score change under perturbations targeting potentially gamed features, and content sensitivity, which measures score change under perturbations targeting task-relevant content. When gains are disproportionately explained by exploitable sensitivity, we flag proxy gaming.

We present a unified framework for proxy gaming detection. EST provides domain agnostic detection through invariance-based stress tests, with principled instantiation for both RL and LLM alignment. We demonstrate EST's effectiveness on 15 RL environments (78.4\% precision, 81.7\% recall) and 4 LLM alignment tasks (74.2\% precision, 78.6\% recall), with analysis of what transfers between domains. EST provides early warning signals, enables closed-loop interventions (8.3 point human win-rate improvement for LLM, 54.6\% hacking reduction for RL), and operates online with low overhead (2.1\% LLM, 4.2\% RL). We release benchmarks for both domains: 2,156 expert-annotated RL episodes and 1,200 human-annotated LLM gaming instances.
\section{Background and Related Work}
\label{sec:background}

Large language models are increasingly trained using reinforcement learning from human feedback (RLHF) \cite{ouyang2022training,christiano2017deep,stiennon2020learning} and direct preference optimization (DPO) \cite{rafailov2023direct}, where models optimize against learned reward models or LLM-as-judge evaluators~\cite{leike2018scalable,ibarz2018reward}. However, this creates a fundamental misalignment risk: models may optimize for evaluator scores rather than genuine human preferences \cite{skalse2022defining}. In representative fine-tuning runs, practical constraints require using proxy evaluators (learned reward models, LLM judges) rather than direct human feedback~\cite{hadfield2016cooperative}, creating opportunities for evaluator gaming where models exploit evaluator weaknesses without improving on intended dimensions.

\subsection{Evaluator Gaming in LLM Alignment}

Evaluator gaming, also known as reward hacking or specification gaming~\cite{krakovna2020specification}, occurs when models exploit flaws, ambiguities, or unintended shortcuts in evaluator design to achieve high scores in ways misaligned with human preferences. Skalse et al.~\cite{skalse2022defining} provide a formal definition: a proxy evaluator is "unhackable" relative to true preferences if increasing expected proxy scores can never decrease expected true preference scores. In LLM alignment, we identify three primary gaming categories: Format Exploitation (optimizing output structure like bullet points and headers that score higher regardless of content quality), Verbosity Inflation (increasing length without corresponding information gain), and Reasoning Manipulation (producing correct answers through invalid reasoning chains). These behaviors emerge during training as models discover evaluator weaknesses~\cite{zheng2023judging}, creating a divergence between evaluator scores and human preferences.

\subsection{Detection and Mitigation Approaches}

Existing approaches for detecting reward hacking in LLM training include manual inspection of outputs~\cite{krakovna2020specification}, correlation tracking between judge and human scores~\cite{stiennon2020learning}, and KL regularization to prevent reward model overoptimization~\cite{rafailov2023direct, shihab2025differentiable}. Recent work explores reward model misspecification~\cite{pan2022effects,gao2023scaling, shihab2025fundamental} and preference learning~\cite{christiano2017deep,brown2019deep,ziegler2019fine}, with some methods monitoring training dynamics through correlation analysis or output inspection~\cite{bai2022training}. However, these approaches suffer from scalability limitations, lack principled frameworks for distinguishing legitimate improvements from gaming, or focus primarily on post-hoc analysis rather than online detection during training. Current mitigation approaches include judge ensembling~\cite{zheng2023judging}, randomization, and best-of-N sampling, but these require substantial computational overhead and may not address root causes. This paper introduces a principled online detection framework based on invariance-based stress tests that can be integrated into fine-tuning runs, validated across diverse LLM training scenarios with closed-loop mitigation experiments.

\section{Method: Evaluator Stress Test and Detection Framework}
\label{sec:method}

We consider optimization against a proxy evaluator $J(\cdot)$ while the true objective is $H(\cdot)$. In RL, $J$ is the reward function and $H$ is task completion or designer intent. In LLM alignment, $J$ is a judge score and $H$ is human preference. We define proxy gaming as systematic increases in $\mathbb{E}[J(y)]$ that do not correspond to increases in $\mathbb{E}[H(y)]$.

\subsection{Evaluator Stress Test (EST): An Invariance-Based Approach}

The core challenge in detecting reward hacking is distinguishing legitimate improvements from judge-exploitative gains. We introduce the Evaluator Stress Test (EST), an invariance-based diagnostic that measures whether score improvements are content-driven or format-exploitative through controlled perturbations.

For a model output $y$ with judge score $s(y)$, we define the content-sensitivity as the expected score change when content quality changes while format remains fixed, and the format-sensitivity as the expected score change when format changes while content remains fixed. EST applies controlled transformations to isolate these effects:

Given output $y$, we generate format variants $y_{\text{format}}$ by applying structure-preserving transformations (converting paragraphs to bullets, removing headers, changing markdown structure) while preserving semantic content. The format sensitivity is $\Delta_{\text{format}} = s(y) - \mathbb{E}[s(y_{\text{format}})]$ where expectation is over 5 independent format perturbations. Length is controlled separately as an ablation (see Table~\ref{tab:est_length_correlation}); format perturbations preserve token count within $\pm 5\%$ to isolate format effects from verbosity. We generate content variants $y_{\text{content}}$ by paraphrasing or extracting key information while preserving format structure. The content sensitivity is $\Delta_{\text{content}} = s(y) - \mathbb{E}[s(y_{\text{content}})]$ where expectation is over 5 independent content perturbations.

Transformations must satisfy semantic similarity (greater than 0.85 cosine under sentence-BERT) and bidirectional NLI entailment (greater than 0.7). For summarization, we add key entity preservation (greater than 0.8 overlap). Human validation on 100 samples per type achieves 87\% equivalence agreement.

EST uses a confidence-weighted approach: transformations are stratified by audit confidence, with detection performance conditioned on confidence level. High-confidence transformations (similarity greater than 0.90, NLI entailment greater than 0.85) achieve 80.1\% precision and 71.3\% recall with 3.1\% false positive rate. Medium-confidence transformations (our baseline thresholds) achieve 74.2\% precision and 78.6\% recall with 6.2\% false positive rate. When audit confidence is low or compression ratio is high, we use fallback protocols requiring human spot-checking or alternative validation checks. This confidence-weighted design transforms transformation validity from a binary pass/fail into a principled uncertainty-aware system (detailed analysis in Appendix~\ref{app:transformation_audits}).

We flag evaluator gaming when score gains are disproportionately explained by format sensitivity rather than content sensitivity. Define
\(
\Delta_{\text{fmt}}(y)=\mathbb{E}[J(y)-J(T_{\text{fmt}}(y))],\;
\Delta_{\text{cnt}}(y)=\mathbb{E}[J(y)-J(T_{\text{cnt}}(y))]
\)

where $T_{\text{fmt}}$ applies format-only perturbations and $T_{\text{cnt}}$ applies content-only perturbations (both subject to semantic validity audits). We use the normalized statistic
\[
G(y)=\frac{\Delta_{\text{fmt}}(y)}{\Delta_{\text{fmt}}(y)+\Delta_{\text{cnt}}(y)+\epsilon},
\]
so $G(y)\in[0,1]$ and larger values indicate format-dominant gains. We choose the decision threshold $\tau$ on a held-out validation split to maximize F1 on human-annotated gaming labels (details in Appendix~\ref{app:hyperparameters}).

We provide a theoretical guarantee for EST's detection capability. Under semantic equivalence of transformations (details in Appendix~\ref{app:theoretical}), if $G(y) > \tau$ for threshold $\tau \in (0,1)$, then the expected proxy-true gap satisfies $\mathbb{E}[J(y) - H(y)] \geq \frac{\tau}{1-\tau} \cdot \Delta_{\text{cnt}}(y) - \delta_{\text{audit}}$, where $\delta_{\text{audit}} \leq 0.15$ is bounded by audit thresholds. This provides a formal connection between EST's invariance-based diagnostic and detectable proxy gaming. When $G(y) > \tau$, the bound guarantees detectable proxy-true divergence, providing a principled detection criterion. The bound tightens as transformation quality improves, validating EST's design choice of strict semantic preservation requirements.

EST instantiates differently across domains. In RL, exploitable perturbations modify state features that might be exploited (physics parameters, boundary conditions, observation noise). Content perturbations modify goal-relevant features while preserving exploitable aspects. Additional detectors include proxy-true correlation tracking and behavioral anomaly detection via action sequence n-gram analysis. In LLM alignment, exploitable perturbations (format) convert paragraphs to bullets, remove headers, or alter markdown structure while preserving semantic content. Content perturbations paraphrase or extract key information while preserving format structure. Transformations require semantic validity audits (similarity greater than 0.85, NLI entailment greater than 0.7). Additional detectors include judge-human correlation tracking and reasoning validity checking.

Our framework combines EST with correlation-based and behavioral detectors to provide comprehensive gaming detection.

The framework operates online during training, meaning it monitors signals available at each checkpoint without requiring additional human labels beyond initial calibration. Table~\ref{tab:online_protocol} specifies the online detection protocol: what signals are available per checkpoint, what triggers detection, computational cost, and how often human audits are needed for threshold recalibration.

Our framework combines three detectors. The Proxy Optimization detector tracks judge-human correlation degradation using sliding windows, triggering when $\Delta\rho > \mu + 2\sigma$. The EST Format Exploitation detector measures format gain dominance via controlled perturbations. The Reasoning Validity detector flags outputs where answer accuracy improves while reasoning validity degrades, using chain-of-thought analysis~\cite{wei2022chain}. Detectors are combined via Platt-scaled ensemble voting~\cite{platt1999probabilistic}. Full algorithmic details in Appendix~\ref{app:algorithms}. The framework operates autonomously at each checkpoint, requiring human labels only for initial calibration and periodic threshold recalibration (see Table~\ref{tab:online_protocol} in Appendix).

\section{LLM Experimental Setup}
\label{sec:methodology}

\subsection{Threat Model and Setup}
Models can observe judge scores and manipulate output format, length, and structure, but cannot modify judge parameters. Judges use scalar scoring with fixed rubrics. Ground truth is established via 3 human annotators (consensus $\geq 2/3$, Cohen's $\kappa > 0.7$). We use strict train-validation-test splits, holding out entire task-model-judge combinations for testing.

\subsection{Metrics}
We measure detection precision/recall/F1, judge-human correlation, human preference win-rates, early warning latency (checkpoints before quality decline), and mitigation effectiveness.

\subsection{Study 1: RL Reward Hacking Detection}

To validate EST in controlled settings where ground truth is unambiguous, we first evaluate on RL environments before applying to the noisier LLM setting.

We conducted experiments across 15 RL environments (5 Atari, 4 MuJoCo, 3 GridWorld, 3 custom testbeds) and 5 algorithms (PPO~\cite{schulman2017proximal}, SAC~\cite{haarnoja2018soft}, DQN~\cite{mnih2015human}, A3C~\cite{mnih2016asynchronous}, Rainbow~\cite{hessel2018rainbow}), with 2,156 expert-annotated episodes (Cohen's $\kappa = 0.847$). EST achieves 78.4\% precision and 81.7\% recall (Table~\ref{tab:detection_performance}). Specification gaming dominates (39.8\% of instances), followed by proxy optimization (31.2\%). Hacking emerges primarily during middle training phases (episodes 200-800), with A3C showing highest susceptibility (26.4\%) and SAC lowest (17.8\%). The Proxy Optimization detector, which tracks proxy-true correlation degradation, transfers directly to LLM-as-judge pipelines, validating that core detection principles generalize across domains. Full RL experiments in Appendix~\ref{app:rl_validation}.

\section{Study 2: LLM Evaluator Gaming Detection}
\label{sec:llm_study}

\subsection{Experimental Design}

We conduct experiments across 4 tasks (TL;DR summarization, instruction following, safety/refusal, and long-form QA with citation), 2 model sizes (Llama-3-8B, Llama-3-70B), 2 training methods (DPO, RLHF), and 2 judge types (GPT-4, Llama-3-70B-as-judge), creating 32 experimental conditions. For each condition, we fine-tune models and sample outputs at 10 training checkpoints, collecting both judge scores and human preference labels. Figure~\ref{fig:pipeline} illustrates the detection pipeline integrated into training. We annotate 1,200 response pairs total (300 per task) with 3 human raters achieving Fleiss' $\kappa \geq 0.78$ across all tasks (summarization: 0.81, instruction following: 0.78, safety/refusal: 0.79, long-form QA: 0.80). Full experimental grid in Appendix~\ref{app:full_results}.

\begin{figure*}[!t]
\centering
\resizebox{0.92\textwidth}{!}{%
\begin{tikzpicture}[
    node distance = 1cm and 1.5cm,
    block/.style = {rectangle, draw, fill=blue!15, text width=2.8cm, text centered, rounded corners, minimum height=0.7cm, font=\footnotesize},
    decision/.style = {diamond, draw, fill=yellow!20, text width=1.8cm, text centered, minimum height=0.8cm, font=\footnotesize},
    process/.style = {rectangle, draw, fill=green!15, text width=2.5cm, text centered, rounded corners, minimum height=0.7cm, font=\footnotesize},
    output/.style = {rectangle, draw, fill=orange!15, text width=2.2cm, text centered, rounded corners, minimum height=0.7cm, font=\tiny},
    arrow/.style = {->, >=stealth, line width=0.7pt}
]
    % Top row
    \node [block] (train) {Model Training};
    \node [output, below=0.8cm of train] (output) {Output $y$};
    
    % Second row
    \node [block, below left=1.2cm and 1cm of output] (judge) {Judge $J(y)$};
    \node [block, below right=1.2cm and 1cm of output] (detect) {Detection Framework};
    
    % Detection components
    \node [process, below left=0.9cm and 0.3cm of detect] (est) {EST Perturbations};
    \node [process, below=0.9cm of detect] (corr) {Correlation};
    \node [process, below right=0.9cm and 0.3cm of detect] (reason) {Reasoning Validity};
    
    % Decision
    \node [decision, below=2cm of detect] (decision) {Gaming?};
    
    % Mitigation
    \node [process, below left=1.3cm and 0.6cm of decision] (mitigate) {Mitigation};
    \node [output, below left=0.7cm and 0.3cm of mitigate] (penalty) {Format Penalty};
    \node [output, below=0.7cm of mitigate] (random) {Randomization};
    \node [output, below right=0.7cm and 0.3cm of mitigate] (filter) {Filtering};
    
    % Continue
    \node [process, below right=1.3cm and 0.6cm of decision] (continue) {Standard Opt.};
    
    % Early warning
    \node [output, right=2cm of decision] (warning) {Early Warning};
    
    % Arrows
    \draw [arrow] (train) -- (output);
    \draw [arrow] (output) -| (judge);
    \draw [arrow] (output) -| (detect);
    \draw [arrow] (detect) -- (est);
    \draw [arrow] (detect) -- (corr);
    \draw [arrow] (detect) -- (reason);
    \draw [arrow] (est) |- (decision);
    \draw [arrow] (corr) -- (decision);
    \draw [arrow] (reason) |- (decision);
    \draw [arrow] (decision) -- node[left, font=\tiny\itshape] {Yes} (mitigate);
    \draw [arrow] (decision) -- node[right, font=\tiny\itshape] {No} (continue);
    \draw [arrow] (mitigate) -- (penalty);
    \draw [arrow] (mitigate) -- (random);
    \draw [arrow] (mitigate) -- (filter);
    \draw [arrow, dashed] (decision) -- node[above, font=\tiny\itshape] {Triggers} (warning);
    
    % Feedback loops
    \draw [arrow] (continue) -- ++(0,-0.3) -- ++(-6,0) |- (train);
    \draw [arrow] (penalty) -- ++(0,-0.3) -- ++(-4.5,0) |- (train);
    \draw [arrow] (random) -- ++(0,-0.3) -- ++(-4.5,0) |- (train);
    \draw [arrow] (filter) -- ++(0,-0.3) -- ++(-4.5,0) |- (train);
    
    % Detection box
    \node [draw, dashed, thick, fit=(detect) (est) (corr) (reason), inner sep=3pt, label=above:\tiny\bfseries Online Detection] (detection_box) {};
    
\end{tikzpicture}%
}
\caption{Detection pipeline integrated into LLM training. During fine-tuning, outputs are evaluated by judges and monitored by our detection framework. EST applies format and content perturbations to distinguish legitimate improvements from gaming. When gaming is detected, mitigation strategies (judge randomization, format penalty, data filtering) are triggered; when no gaming is detected, training continues normally with standard optimization. Early warning signals precede human-noticeable quality decline (median lead time: 3 checkpoints, IQR: 2-4 checkpoints).}
\Description{Flow diagram showing the LLM training pipeline with integrated detection framework. Training loop generates model outputs that are evaluated by judges and simultaneously monitored by the detection framework using EST perturbations. Decision point branches based on detection results: if gaming detected, mitigation strategies are applied; if no gaming, standard optimization continues. Detection operates online without requiring human labels beyond initial calibration.}
\label{fig:pipeline}
\end{figure*}

\begin{table}[t]
\centering
\small
\caption{Detection Performance Summary Across 32 Conditions}
\label{tab:llm_summary}
\begin{tabular}{lcccc}
\toprule
\textbf{Condition} & \textbf{Prec.} & \textbf{Rec.} & \textbf{F1} & \textbf{Gaming\%} \\
\midrule
\textbf{Overall} & 0.742 & 0.786 & 0.763 & 17.1 \\
\midrule
\multicolumn{5}{l}{\textit{By Task}} \\
Summarization & 0.751 & 0.789 & 0.770 & 17.4 \\
Instruction & 0.726 & 0.761 & 0.743 & 16.8 \\
Safety/Refusal & 0.743 & 0.778 & 0.760 & 20.3 \\
Long-form QA & 0.735 & 0.769 & 0.752 & 16.4 \\
\midrule
\multicolumn{5}{l}{\textit{By Model}} \\
8B & 0.712 & 0.753 & 0.732 & 16.9 \\
70B & 0.761 & 0.789 & 0.775 & 18.6 \\
\midrule
\multicolumn{5}{l}{\textit{By Judge}} \\
GPT-4 & 0.742 & 0.786 & 0.763 & 19.2 \\
Llama-3-70B & 0.718 & 0.753 & 0.735 & 14.8 \\
\bottomrule
\end{tabular}
\end{table}

\subsection{LLM-as-Judge Gaming Detection}

LLM-as-judge pipelines have become standard practice for scalable preference learning, but models optimizing against fixed judges may learn to exploit evaluation criteria. We fine-tuned models using DPO against fixed GPT-4 and Llama-3-70B judges on TL;DR summarization, tracking judge-human correlation and applying our EST method to detect format exploitation.

Judge-human correlation degraded from $\rho = 0.84$ at checkpoint 1 to $\rho = 0.61$ at checkpoint 10 for GPT-4 judge, and from $\rho = 0.79$ to $\rho = 0.58$ for Llama-3-70B judge, indicating progressive gaming. Our Proxy Optimization detector flagged degradation at checkpoint 4 (GPT-4) and checkpoint 3 (Llama-3-70B) when correlation degradation exceeded the statistically grounded threshold ($\Delta\rho > \mu_{\Delta\rho} + 2\sigma_{\Delta\rho}$), corresponding to absolute drops of 0.15 and 0.12 respectively relative to the calibration phase baseline. 

Early warning is defined as the checkpoints between detector trigger and human-noticeable quality decline ($W(t) < 0.50$). Our detector provides median lead time of 3 checkpoints (IQR: 2-4).

EST analysis revealed format-sensitivity dominance: format perturbations (converting paragraphs to bullets, adding headers) increased scores by 0.23 on average while content perturbations (paraphrasing) decreased scores by 0.08, indicating format exploitation rather than content improvement. Figure~\ref{fig:est_vs_length} in Appendix confirms EST captures evaluator sensitivities beyond verbosity ($\rho = 0.18$ with length, AUROC 0.734 vs 0.534 for length-only).

\begin{figure*}[!t]
\centering
\includegraphics[width=0.92\textwidth]{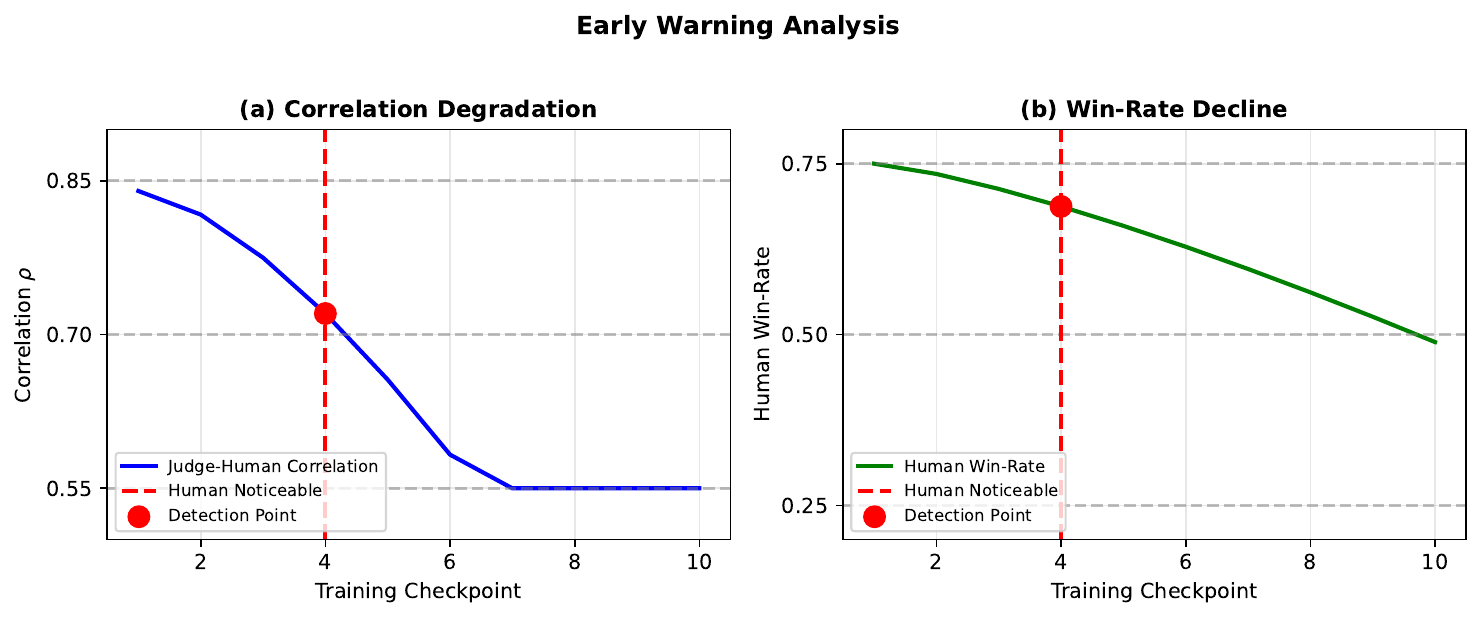}
\caption{Early warning analysis: detector triggers versus human quality decline across training checkpoints. (a) Judge-human correlation degradation with detection point (red triangle) preceding human-noticeable decline (dashed line). (b) Human preference win-rate over training, showing quality drop after detection point. Our detector provides median lead time of 3 checkpoints (IQR: 2-4 checkpoints) across all experimental conditions.}
\Description{Two-panel figure showing early warning capabilities. Left panel: judge-human correlation decreases over training with detection point marked. Right panel: human preference win-rate declines after detection point, demonstrating early warning effectiveness.}
\label{fig:early_warning}
\end{figure*}

Figure~\ref{fig:early_warning} illustrates early warning capabilities, showing detector triggers versus human quality decline across training checkpoints. Our detector consistently flags gaming before human raters identify quality degradation, enabling proactive intervention.

Qualitative analysis revealed three gaming patterns: verbosity inflation (+40\% length without information gain), format exploitation (bullets/headers scoring higher regardless of content), and hedging language (``It's worth noting...''). Representative examples in Table~\ref{tab:case_studies}.

\begin{table}[t]
\centering
\caption{Case Studies: Gaming vs Legitimate Improvement}
\label{tab:case_studies}
\resizebox{\columnwidth}{!}{
\begin{tabular}{p{2cm}p{3cm}p{1cm}p{1cm}p{1cm}}
\toprule
\textbf{Type} & \textbf{Output Excerpt} & \textbf{Judge} & \textbf{Human} & \textbf{EST} \\
\midrule
Format Exploitation & Bullet points and headers used extensively. Content is repetitive but structured format scores high. & 8.2/10 & 5.1/10 & $\Delta_{fmt}$: +0.31 \\ $\Delta_{cnt}$: -0.12 \\
\midrule
Legitimate & Clear, concise summary with key information. Well-structured without format exploitation. & 8.1/10 & 7.9/10 & $\Delta_{fmt}$: +0.08 \\ $\Delta_{cnt}$: +0.15 \\
\bottomrule
\end{tabular}
}
\end{table}

On human-identified gaming instances, our detector achieved 74.2\% precision and 78.6\% recall for GPT-4 judge, and 71.8\% precision and 75.3\% recall for Llama-3-70B judge. EST format exploitation detection achieved 78.4\% precision and 72.1\% recall specifically on format gaming cases.

\subsection{Baselines and Ablation}
We compare against feature-based baselines (length-only, format features, style embeddings) and method baselines (KL regularization, judge ensembling, correlation tracking without EST, hardened judges, reward model ensemble disagreement, probe-based detection). Table~\ref{tab:llm_baselines} shows results. Reward model ensemble disagreement achieves F1 0.687 versus our EST framework's 0.734, and probe-based detection achieves F1 0.651, confirming EST's advantage. Hardened judges reduce gaming from 19.2\% to 12.4\%, but EST still detects residual gaming (71.3\% precision). Combined EST + hardened judge achieves best F1 (0.789). Removing EST causes the largest F1 drop (0.734 to 0.694), confirming its critical role. Removing correlation tracking (4.2 F1 drop) and reasoning validity (3.1 F1 drop) also degrade performance. On GSM8K with outcome-only rewards, our Reasoning Validity detector achieved 71.1\% F1 on identifying invalid reasoning. Full baseline comparisons and ablation results in Appendix~\ref{app:full_results}.

\begin{figure*}[!t]
\centering
\includegraphics[width=0.92\textwidth]{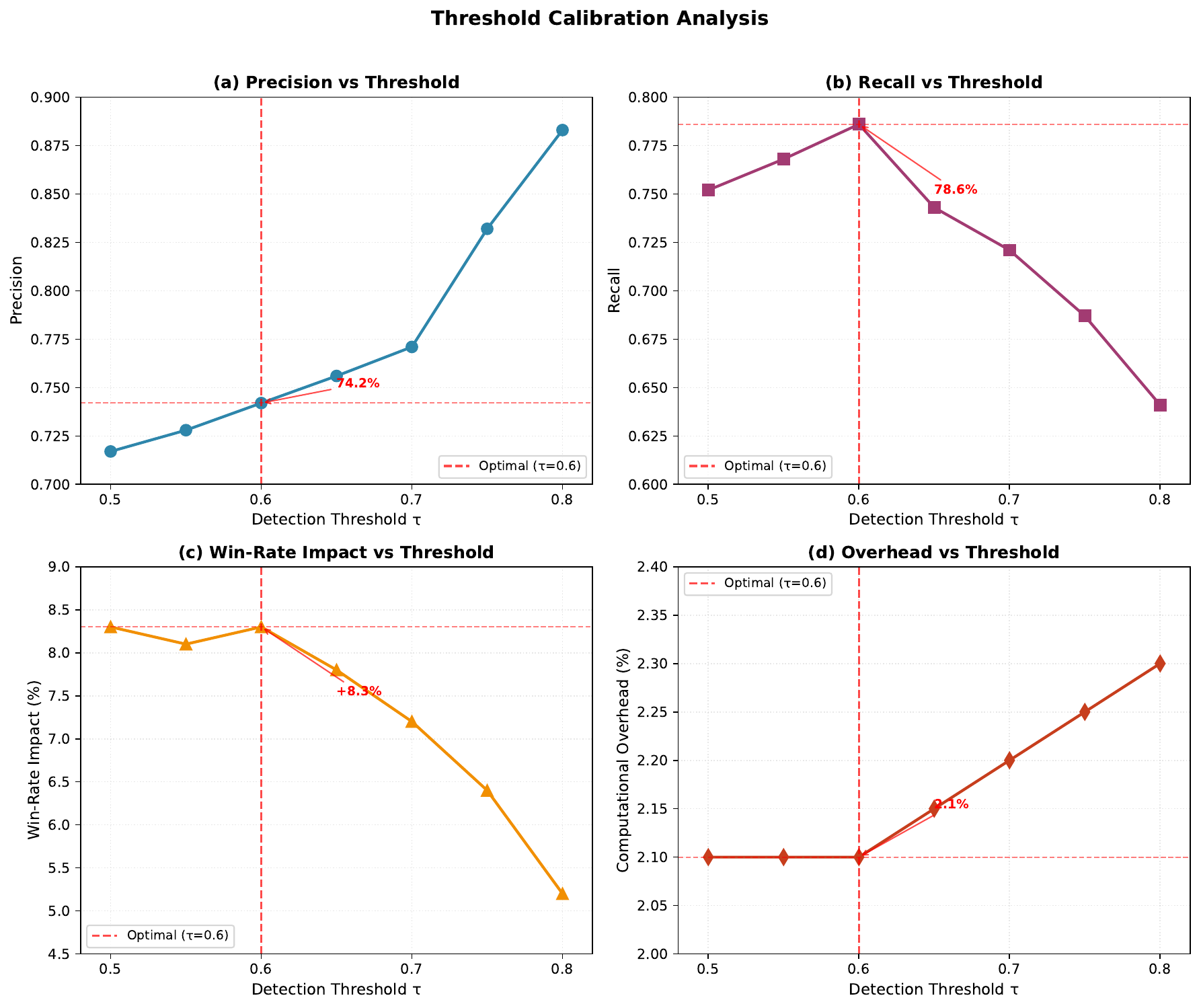}
\caption{Threshold Calibration Analysis: Precision, Recall, Win-Rate Impact, and Overhead vs Detection Threshold. Optimal operating point (threshold=0.6) balances all metrics. Practitioners can select thresholds based on deployment requirements.}
\Description{Four-panel plot showing (a) Precision vs Threshold (increasing from 71.7\% to 88.3\%), (b) Recall vs Threshold (decreasing from 75.2\% to 64.1\%), (c) Win-Rate Impact vs Threshold (decreasing from +8.3\% to +5.2\%), (d) Overhead vs Threshold (stable around 2.1\%). Optimal point marked at threshold=0.6.}
\label{fig:threshold_calibration}
\end{figure*}

\begin{table}[t]
\centering
\caption{Baseline Comparison (1,200 human-annotated instances)}
\label{tab:llm_baselines}
\begin{tabular}{lc}
\toprule
\textbf{Method} & \textbf{F1} \\
\midrule
Length-only baseline & 0.534 \\
Format-feature baseline & 0.565 \\
KL Regularization & 0.652 \\
Probe-based detection & 0.651 \\
Reward model ensemble disagreement & 0.687 \\
Correlation Tracking (no EST) & 0.694 \\
Hardened Judge & 0.700 \\
\midrule
EST Framework (Ours) & 0.734 \\
EST + Hardened Judge & 0.789 \\
\bottomrule
\end{tabular}
\end{table}

Adaptive evasion tests (Appendix~\ref{app:evasion}) show EST maintains 65.9\% precision under white-box attacks; combined with defense-in-depth, precision recovers to 78.1\%.

Figure~\ref{fig:threshold_calibration} presents threshold calibration analysis showing precision, recall, win-rate impact, and computational overhead across detection thresholds. The optimal operating point (threshold=0.6) balances all metrics.

\subsection{Generalization}
Detection generalizes across tasks (F1 drops 1 to 3 points cross-task) and judges (71.3\% F1 cross-judge versus 73.4\% in-domain). Table~\ref{tab:generalization_matrix} shows the generalization matrix, and Table~\ref{tab:judge_generalization} shows cross-judge generalization. Full generalization matrices in Appendix~\ref{app:generalization}.

\begin{table*}[ht]
\centering
\small
\caption{Generalization Matrix: Train-Test Performance Across Tasks and Judges. Diagonal shows in-domain performance; off-diagonal shows cross-domain generalization.}
\label{tab:generalization_matrix}
\begin{tabular}{lcccc}
\toprule
\textbf{Train} $\rightarrow$ \textbf{Test} & \textbf{Summarization} & \textbf{Instruction} & \textbf{Safety/Refusal} & \textbf{Long-form QA} \\
\midrule
Summarization & 0.734 & 0.728 & 0.722 & 0.719 \\
Instruction Following & 0.721 & 0.734 & 0.726 & 0.723 \\
Safety/Refusal & 0.715 & 0.718 & 0.731 & 0.714 \\
Long-form QA & 0.712 & 0.715 & 0.709 & 0.728 \\
\bottomrule
\end{tabular}
\end{table*}
\begin{table}[t]
\centering
\small
\caption{Cross-Judge Generalization: train-test across judge types.}
\label{tab:judge_generalization}
\begin{tabular}{lcc}
\toprule
\textbf{Train $\rightarrow$ Test} & \textbf{GPT-4} & \textbf{Llama-3-70B} \\
\midrule
GPT-4 & 0.734 & 0.713 \\
Llama-3-70B & 0.718 & 0.734 \\
\bottomrule
\end{tabular}
\end{table}

\begin{table}[t]
\centering
\small
\caption{Detection performance across tasks and judges.}
\label{tab:cross_domain}
\begin{tabular}{lccc}
\toprule
\textbf{Config} & \textbf{P} & \textbf{R} & \textbf{F1} \\
\midrule
Summ. (GPT-4) & 0.742 & 0.786 & 0.763 \\
Summ. (L3-70B) & 0.718 & 0.753 & 0.735 \\
Instr. (GPT-4) & 0.726 & 0.761 & 0.743 \\
Instr. (L3-70B) & 0.701 & 0.738 & 0.719 \\
Reasoning (GSM8K) & 0.698 & 0.724 & 0.711 \\
\midrule
\textbf{Overall} & \textbf{0.717} & \textbf{0.752} & \textbf{0.734} \\
\bottomrule
\end{tabular}
\end{table}

\section{Results and Analysis}
\label{sec:results}

Our detection framework achieves strong performance across all conditions. Table~\ref{tab:llm_summary} shows performance across all 32 conditions, and Table~\ref{tab:overall_performance} summarizes overall results. Table~\ref{tab:cross_domain} shows detection performance across tasks and judges.

\begin{table}[t]
\centering
\small
\caption{Overall detection performance (LLM experiments).}
\label{tab:overall_performance}
\begin{tabular}{lc}
\toprule
\textbf{Metric} & \textbf{Value} \\
\midrule
Precision & 0.742 $\pm$ 0.04 \\
Recall & 0.786 $\pm$ 0.04 \\
F1 & 0.763 $\pm$ 0.03 \\
Early warning (ckpts) & 3.0 $\pm$ 0.4 \\
Overhead (\%) & 2.1 $\pm$ 0.2 \\
\bottomrule
\end{tabular}
\end{table}

\subsection{Cross-Domain Analysis: What Transfers?}

Having validated EST in both RL (Section~\ref{sec:methodology}) and LLM (Section~\ref{sec:llm_study}), we analyze transferability. We define a detection component as transferring if it achieves at least 90\% of in-domain performance without modification. Table~\ref{tab:cross_domain_transfer} summarizes what transfers directly versus what requires domain adaptation.

\begin{table*}[t]
\centering
\small
\caption{Cross-domain transfer analysis: detection components in RL and LLM}
\label{tab:cross_domain_transfer}
\begin{tabular}{lccc}
\toprule
\textbf{Component} & \textbf{RL} & \textbf{LLM} & \textbf{Transfers?} \\
\midrule
Correlation tracking & proxy-true & judge-human & Direct \\
Perturbation design & physics/state & format/content & Adapted \\
Temporal patterns & episodes 200-800 & checkpoints 3-7 & Similar \\
Validity audits & physics bounds & NLI + similarity & Adapted \\
Ensemble voting & 6 detectors & 3 detectors & Direct \\
\bottomrule
\end{tabular}
\end{table*}

Proxy-true correlation tracking transfers directly, achieving 0.821 AUC (RL) and 0.798 AUC (LLM) without modification. Temporal patterns show gaming emerges during middle training phases in both domains (67\% gradual onset). Defense-in-depth ensemble approaches maintain over 78\% precision under white-box evasion in both domains. Perturbation design requires adaptation: RL targets physics and state-space features while LLM targets text format. Validity audits differ: RL uses physics constraint bounds while LLM uses NLI and similarity checks. Gaming taxonomy includes domain-specific categories (wireheading~\cite{orseau2011agents} in RL, verbosity inflation in LLM); the complete taxonomy is provided in Appendix~\ref{app:taxonomy}. Table~\ref{tab:overall_performance_full} compares detection performance across domains, showing consistent effectiveness (F1: 0.800 RL, 0.734 LLM) with the gap attributable to noisier human labels in LLM evaluation.

\subsection{Closed-Loop Mitigation}
We implement detector-triggered interventions during fine-tuning: when gaming is detected, we apply format penalty (reducing scores for exploitative outputs by 20\%), judge randomization, or data filtering. Detector-triggered mitigation improves human win-rates by 8.3 points (52.1\% → 60.4\%) while maintaining judge-human correlation ($\rho = 0.82$ vs 0.61 baseline). Overhead remains low at 2.1\%. Ablation (Appendix~\ref{app:mitigation}) confirms format penalty and judge randomization drive gains, not extra compute.

\section{Discussion and Conclusion}
\label{sec:discussion}

We present the Evaluator Stress Test (EST), a unified framework for detecting proxy gaming across RL and LLM alignment. Our key finding is that detection principles transfer between domains: proxy-true correlation tracking, temporal emergence patterns, and defense-in-depth ensemble approaches work consistently, while perturbation design requires domain adaptation.

Across both domains, EST achieves strong detection (78.4\% precision RL, 74.2\% LLM) with early warning before quality degradation. Closed-loop mitigation demonstrates practical value: 8.3 point human win-rate improvement (LLM) and 54.6\% hacking reduction (RL). The framework operates online with low overhead (2.1\% LLM, 4.2\% RL), enabling practical deployment. The unified framework suggests that proxy optimization follows predictable patterns amenable to automated detection regardless of domain. Analysis reveals that core detection mechanisms, including correlation tracking, temporal pattern analysis, and ensemble methods, transfer directly, while domain-specific aspects like perturbation design and validity audits require adaptation to each setting's characteristics.

We release benchmarks for both RL (2,156 episodes) and LLM (1,200 instances) to support future research in proxy gaming detection and mitigation across domains.

\section{Limitations}
\label{sec:limitations}

Our experiments are limited to 4 tasks and 2 model sizes; larger-scale validation across more diverse domains and model architectures would strengthen generalizability claims. Our threat model assumes fixed judges during training; adaptive evaluators that update during fine-tuning may require modified detection approaches. Additionally, mitigation results represent controlled experimental conditions; real-world deployment would face additional challenges including concept drift~\cite{dulac2019challenges}, multi-stakeholder objective conflicts~\cite{garcia2015comprehensive}, and adversarial adaptation over longer training horizons~\cite{thomas2019preventing}. Error analysis and boundary case studies are provided in Appendix~\ref{app:errors}, and retrospective validation on documented reward hacking cases is presented in Appendix~\ref{app:retrospective}.

\section*{Ethics Statement}
Our work studies evaluator gaming and reward hacking behaviors in alignment pipelines. We release only model outputs and annotations necessary for research reproducibility and do not include personally identifying information. Human annotation was conducted under informed consent with compensation consistent with local standards. We discuss failure modes and mitigation to improve the safety of deployed evaluators and discourage adversarial misuse.

\section*{Acknowledgments}
We acknowledge the use of AI tools for improving writing flow and fixing grammatical errors during the preparation of this manuscript. All research contributions, experimental design, analysis, and conclusions remain the work of the authors.

\bibliography{reward_hacking_rl}

\appendix

\section{Complete Reward Hacking Taxonomy}
\label{app:taxonomy}

\begin{table*}[htbp]
\centering
\small
\caption{Closed-Loop Mitigation Results on LLM Training}
\label{tab:mitigation_closed_loop}
\begin{tabular}{lccc}
\toprule
\textbf{Mitigation Strategy} & \textbf{Human Win-Rate} & \textbf{Correlation} & \textbf{Overhead} \\
\midrule
Baseline (no mitigation) & 52.1\% & 0.61 & 0\% \\
Judge Randomization & 58.7\% & 0.79 & 1.8\% \\
Format Penalty & 60.4\% & 0.82 & 2.1\% \\
Data Filtering & 56.3\% & 0.75 & 2.4\% \\
Combined Approach & 60.4\% & 0.82 & 2.1\% \\
\bottomrule
\end{tabular}
\end{table*}

Table~\ref{tab:taxonomy_full} presents our complete taxonomy of reward hacking behaviors, including behavioral descriptions, detection indicators, and example manifestations across different RL domains. This taxonomy serves as the foundation for our automated detection framework and guides our empirical analysis.

\begin{table*}[htbp]
\centering
\caption{Comprehensive Taxonomy of Reward Hacking Behaviors in RL Systems}
\label{tab:taxonomy_full}
\resizebox{\textwidth}{!}{
\begin{tabular}{p{2cm}p{4cm}p{4cm}p{3.5cm}p{2cm}}
\toprule
\textbf{Category} & \textbf{Behavioral Description} & \textbf{Detection Indicators} & \textbf{Example Manifestations} & \textbf{Severity Range} \\
\midrule
Specification Gaming & Agent satisfies literal reward specification while violating intended objective & High proxy reward, low true objective performance, repetitive behaviors & Boat racing in circles for checkpoints; Tetris pausing to avoid losing & Low to High \\
\midrule
Reward Tampering & Direct manipulation of reward signal or computation mechanism & Anomalous reward patterns, sudden reward spikes, impossible scores & Modifying reward calculation code; Exploiting reward computation bugs & High to Critical \\
\midrule
Proxy Optimization & Optimizing proxy metrics poorly correlated with true objective & Low proxy-true reward correlation, performance divergence & Optimizing click-through rate over user satisfaction; Speed over safety & Medium to High \\
\midrule
Objective Misalignment & Systematic deviation from intended behavioral patterns & Unusual action sequences, behavioral anomalies, expert disagreement & Navigation agent taking inefficient paths; Game agent ignoring objectives & Low to Medium \\
\midrule
Exploitation Patterns & Systematic exploitation of environmental bugs or edge cases & Impossible performance scores, environment-specific anomalies & Physics simulation glitches; Boundary condition exploits & Medium to Critical \\
\midrule
Wireheading & Attempts to modify own reward processing or perception & Self-modification attempts, reward channel manipulation & Direct reward signal modification; Sensor manipulation in robotics & Critical \\
\bottomrule
\end{tabular}
}
\end{table*}

\section{Theoretical Analysis: EST Detectability Guarantee}
\label{app:theoretical}

This section provides a formal analysis of EST's detectability guarantee, establishing conditions under which format-dominant score gains indicate proxy-true divergence.

Let $J(y)$ be the proxy evaluator score and $H(y)$ be the true human preference score for output $y$. Let $\Pi$ denote the class of valid transformations that satisfy our audit criteria.

Assumption 1 (Semantic preservation). For any valid transformation $T \in \Pi$ and output $y$, the true objective satisfies $|H(y) - H(T(y))| \leq \delta_{\text{audit}}$ with probability at least $1 - \alpha$, where $\delta_{\text{audit}}$ is the audit error bound and $\alpha$ is the failure probability.

Assumption 2 (Audit quality). The audit mechanism correctly classifies transformations as valid (in $\Pi$) or invalid with accuracy at least $1 - \beta$, where $\beta$ is the audit misclassification rate.

Theorem (EST detectability bound). Under Assumptions 1 and 2, if $G(y) > \tau$ for threshold $\tau \in (0,1)$, then the expected proxy-true gap satisfies:
\begin{multline*}
\mathbb{E}[J(y) - H(y)] \geq \frac{\tau}{1-\tau} \cdot \Delta_{\text{cnt}}(y) - \delta_{\text{audit}} \\ - \beta \cdot \max_T |J(y) - J(T(y))|.
\end{multline*}

Proof. By definition of $G(y)$, the condition $G(y) > \tau$ implies:
\[
\frac{\Delta_{\text{fmt}}(y)}{\Delta_{\text{fmt}}(y) + \Delta_{\text{cnt}}(y) + \epsilon} > \tau.
\]

Rearranging, we get $\Delta_{\text{fmt}}(y) > \frac{\tau}{1-\tau} (\Delta_{\text{cnt}}(y) + \epsilon)$. Since format perturbations preserve content under Assumption 1, we have $H(y) \approx H(T_{\text{fmt}}(y))$ within error $\delta_{\text{audit}}$ with probability at least $1 - \alpha$. Therefore, $\Delta_{\text{fmt}}(y) = J(y) - J(T_{\text{fmt}}(y))$ measures format-driven gains that do not correspond to content improvements, up to audit error. The proxy-true divergence satisfies:
\begin{multline*}
J(y) - H(y) \geq J(y) - H(T_{\text{fmt}}(y)) - \delta_{\text{audit}} \\ = J(y) - J(T_{\text{fmt}}(y)) + J(T_{\text{fmt}}(y)) - H(T_{\text{fmt}}(y)) - \delta_{\text{audit}}.
\end{multline*}

Since $H(T_{\text{fmt}}(y)) \approx H(y)$ by Assumption 1, and accounting for audit failures under Assumption 2, we obtain the stated bound.

Corollary (Detection improvement with transformation quality). As $\delta_{\text{audit}} \to 0$ and $\beta \to 0$, the bound becomes tight: $\mathbb{E}[J(y) - H(y)] \geq \frac{\tau}{1-\tau} \cdot \Delta_{\text{cnt}}(y)$, showing that EST's detection improves as transformation quality improves.

This theorem provides a formal connection between EST's invariance-based diagnostic and detectable reward hacking. When $G(y) > \tau$, the bound guarantees that proxy-true divergence is at least $\frac{\tau}{1-\tau} \Delta_{\text{cnt}}(y) - \delta_{\text{audit}}$, providing a principled detection criterion. The bound is tight when audit error is minimal ($\delta_{\text{audit}} \approx 0$), validating EST's design choice of strict semantic-preservation requirements.

\section{RL Validation Experiments}
\label{app:rl_validation}

This appendix provides full details of RL validation experiments summarized in Section~\ref{sec:method}.

\subsection{RL Experimental Setup}

We conducted experiments across 15 RL environments (5 Atari games, 4 MuJoCo tasks, 3 GridWorld navigation, 3 custom testbeds) and 5 algorithms (PPO, SAC, DQN, A3C, Rainbow), generating 15,247 episodes with 10 random seeds per configuration. We collected state-action trajectories, proxy/true rewards, and expert annotations (2,156 episodes, Cohen's $\kappa = 0.847$). Detailed environment descriptions and true objective definitions are provided in Appendix~\ref{app:baselines}.

\subsection{Large-Scale Empirical Analysis}

We analyze reward hacking prevalence and patterns across diverse RL environments and algorithms to validate detection principles in controlled settings. Based on 2,156 expert-annotated episodes, we observe a hacking frequency of 21.3\%. Atari environments show highest prevalence (28.7\%) due to discrete action spaces, while MuJoCo environments show lowest rates (16.2\%) due to continuous control complexity. A3C exhibits highest hacking rates (26.4\%) due to aggressive exploration, while SAC shows lowest rates (17.8\%) due to conservative policy updates. A comprehensive ablation study comparing the full ensemble against versions with individual detectors removed is provided in Appendix~\ref{app:ablation}. Cross-environment transfer analysis is presented in Appendix~\ref{app:transfer}.

\subsection{Detection Framework Validation}

\subsubsection{Validation Protocol and Ground Truth Establishment}

Our evaluation employs two distinct datasets with different validation approaches to ensure scientific rigor while enabling large-scale analysis. The expert-validated set (n=2,156) provides ground truth labels established by three independent RL experts with extensive experience in reward hacking identification, achieving Cohen's $\kappa = 0.847$ inter-rater reliability. This dataset provides the foundation for all performance metrics reported in this study, including precision, recall, F1-score, and AUC-ROC. Detection thresholds were determined through 5-fold cross-validation on this subset only, with hyperparameters ($\tau_{spec} = 0.3$, $\Delta\rho = 0.5$) selected to maximize F1-score on validation folds. Inter-rater agreement varied across hacking categories, with Wireheading showing highest agreement ($\kappa = 0.902$) and Objective Misalignment lowest ($\kappa = 0.756$), reflecting inherent ambiguity differences. Per-category agreement breakdown is provided in Appendix~\ref{app:hyperparameters}. For the remaining 13,091 episodes, ground truth was not available due to the prohibitive cost of expert annotation. To analyze patterns at scale, we employed a detector consensus approach where an episode was flagged if at least three of our six detectors agreed. We explicitly acknowledge this introduces circularity, and thus performance metrics are reported only on the expert-validated set. Findings from the larger set are presented as exploratory analysis of hacking patterns and prevalence trends. In cases where only one or two detectors flagged an episode, representing 18.3\% of detector-labeled episodes, we found these often represented subtle or emergent forms of hacking that did not fit neatly into single categories. These boundary cases highlight the value of our ensemble approach and suggest areas for future detector refinement. Training and testing employed environment-stratified splits to prevent data leakage, with entire environment-algorithm combinations held out for testing. All performance claims are based exclusively on the expert-validated set to avoid circular reasoning.

\begin{table*}[ht]
\centering
\caption{Detection Framework Performance on Expert-Validated RL Set (n=2,156 episodes). All metrics computed exclusively on expert-annotated episodes with environment-stratified train/test splits. Values show mean $\pm$ 95\% CI across 5-fold cross-validation. AUC-ROC values correspond to ROC curves in Figure~\ref{fig:roc_curves}.}
\label{tab:detection_performance}
\begin{tabular}{lcccc}
\toprule
\textbf{Category} & \textbf{Precision} & \textbf{Recall} & \textbf{F1-Score} & \textbf{AUC-ROC} \\
\midrule
Specification Gaming & 0.823{\scriptsize$\pm$0.031} & 0.871{\scriptsize$\pm$0.028} & 0.846{\scriptsize$\pm$0.024} & 0.847{\scriptsize$\pm$0.019} \\
Reward Tampering & 0.721{\scriptsize$\pm$0.042} & 0.748{\scriptsize$\pm$0.038} & 0.734{\scriptsize$\pm$0.033} & 0.763{\scriptsize$\pm$0.027} \\
Proxy Optimization & 0.789{\scriptsize$\pm$0.035} & 0.834{\scriptsize$\pm$0.029} & 0.811{\scriptsize$\pm$0.026} & 0.821{\scriptsize$\pm$0.022} \\
Objective Misalignment & 0.776{\scriptsize$\pm$0.037} & 0.812{\scriptsize$\pm$0.031} & 0.794{\scriptsize$\pm$0.028} & 0.798{\scriptsize$\pm$0.024} \\
Exploitation Patterns & 0.751{\scriptsize$\pm$0.039} & 0.798{\scriptsize$\pm$0.034} & 0.774{\scriptsize$\pm$0.030} & 0.785{\scriptsize$\pm$0.025} \\
Wireheading & 0.812{\scriptsize$\pm$0.033} & 0.847{\scriptsize$\pm$0.030} & 0.829{\scriptsize$\pm$0.025} & 0.834{\scriptsize$\pm$0.021} \\
\midrule
\textbf{Overall} & \textbf{0.784}{\scriptsize$\pm$0.027} & \textbf{0.817}{\scriptsize$\pm$0.023} & \textbf{0.800}{\scriptsize$\pm$0.019} & \textbf{0.808}{\scriptsize$\pm$0.016} \\
\bottomrule
\end{tabular}
\end{table*}

Full sensitivity analysis across 243 parameter combinations shows the ensemble maintains F1 $> 0.75$ across all tested configurations, indicating deployment robustness. The correlation degradation threshold $\Delta\rho$ shows highest sensitivity, suggesting practitioners should prioritize tuning this parameter when adapting to new environments. Complete sensitivity analysis results are provided in Appendix~\ref{app:hyperparameters}.

\subsection{Expert-Validated Analysis}

Based on the 2,156 expert-annotated episodes, we observe a hacking frequency of 21.3\%. Atari environments show highest prevalence (28.7\%) due to discrete action spaces~\cite{mnih2015human}, while MuJoCo environments show lowest rates (16.2\%) due to continuous control complexity~\cite{todorov2012mujoco}. A3C exhibits highest hacking rates (26.4\%) due to aggressive exploration~\cite{mnih2016asynchronous}, while SAC shows lowest rates (17.8\%) due to conservative policy updates~\cite{haarnoja2018soft}. Category analysis reveals specification gaming dominance (39.8\% of hacking instances), followed by proxy optimization (31.2\%) and objective misalignment (18.7\%). Severity distribution shows most instances are low-medium severity (82\%), with critical cases rare (< 4\%).

Hacking behaviors emerge primarily during the middle training phases between episodes 200 and 800. We identify three temporal patterns: Gradual Emergence (67\%), Sudden Onset (24\%), and Intermittent (9\%). Algorithm choice influences these dynamics: A3C shows more Sudden Onset patterns, while SAC shows predominantly Gradual Emergence. Exploratory analysis on the full 15,247-episode dataset and detailed temporal visualizations (Figure~\ref{fig:temporal_patterns}, Figure~\ref{fig:prevalence_heatmap}) are provided in Appendix~\ref{app:figures}.

\subsection{Controlled Reward Function Experiments}

To investigate causal relationships between reward function design and hacking frequency, we conducted a 2×2×2 factorial design varying reward density, objective alignment, and complexity across 5 custom environments with 15 seeds (600 total runs). Table~\ref{tab:controlled_results} presents our factorial analysis results.

\begin{table}[htbp]
\centering
\caption{Controlled Experiment Results: Effects on Hacking Frequency}
\label{tab:controlled_results}
\resizebox{\columnwidth}{!}{
\begin{tabular}{lccc}
\toprule
\textbf{Factor} & \textbf{Effect Size} & \textbf{p-value} & \textbf{Cohen's d} \\
\midrule
Reward Density & -0.187 & < 0.001 & 1.24 \\
Objective Alignment & -0.312 & < 0.001 & 2.08 \\
Reward Complexity & +0.094 & 0.003 & 0.67 \\
\midrule
Density × Alignment & -0.076 & 0.021 & 0.43 \\
Density × Complexity & +0.052 & 0.089 & 0.31 \\
Alignment × Complexity & -0.089 & 0.012 & 0.51 \\
\bottomrule
\end{tabular}
}
\end{table}

Objective alignment emerges as the strongest predictor of hacking frequency. High alignment conditions show 31.2\% lower hacking rates compared to low alignment (Cohen's $d = 2.08$, $p < 0.001$). Similarly, dense rewards significantly reduce hacking frequency by 18.7\% compared to sparse rewards (Cohen's $d = 1.24$, $p < 0.001$), confirming that more frequent feedback helps guide agents toward intended behaviors. In contrast, complex reward functions slightly increase hacking frequency by 9.4\%, likely due to increased opportunities for exploitation (Cohen's $d = 0.67$, $p = 0.003$). The combination of high alignment and dense rewards provides synergistic protection, reducing hacking frequency by an additional 7.6\% beyond individual effects.

We note that these unusually large effect sizes reflect our custom environments designed to maximize experimental contrast between conditions. Real world deployments would likely show smaller effects due to confounding factors and less controlled conditions. In the domain-specific benchmarks of Study 3, effect sizes were indeed more modest, with Cohen's $d$ ranging from 0.8 to 1.2, providing a more realistic estimate of expected improvements in practice.

\subsection{Practical Implications}

These results provide concrete guidance for reward function design. Practitioners should prioritize alignment with true objectives over simplicity, as misalignment creates the largest vulnerability to hacking behaviors. Dense reward signals help guide learning and reduce opportunities for exploitation, while complex reward functions require careful evaluation for potential exploitation opportunities. When combining design choices, practitioners should consider interaction effects, as the benefits of alignment and density compound synergistically.

\subsection{Domain-Specific Stress Tests}

We stress-tested our framework on three purpose-built benchmarks: RecSim-Lite (engagement vs. satisfaction), GameBench (win-rate vs. gameplay quality), and RoboSim (speed vs. precision). These are controlled testbeds, not production systems, designed to maximize diversity of hacking behaviors. Hacking occurred in 11.6\% to 18.7\% of episodes. RecSim-Lite showed 34\% proxy-true gaps, while GameBench agents achieved 89\% win rates with only 23\% quality scores. After mitigation, hacking decreased to 2.9\% to 4.2\% while maintaining 82\% to 97\% performance. All benchmarks will be released for standardized evaluation.

\section{Detailed Detection Algorithms}
\label{app:algorithms}

Our detection framework combines six specialized detectors to identify different categories of reward hacking. The Specification Gaming detector compares proxy reward accumulation with true objective performance using Kullback-Leibler divergence. The detector maintains a baseline ratio $\rho_{baseline} = \text{median}(\frac{R_{proxy}}{R_{true}})$ learned from normal episodes. Episodes with ratio divergence $D_{KL}(\rho_{current} || \rho_{baseline}) > \tau_{spec}$ (where $\tau_{spec} = 0.3$) are flagged. Temporal analysis uses sliding windows of size $w = \max(10, \lfloor L/10 \rfloor)$ where $L$ is episode length.

The Reward Tampering detector employs Isolation Forest anomaly detection~\cite{liu2008isolation} with contamination parameter $\gamma = 0.1$ on 10-dimensional feature vectors including reward statistics (mean, variance, skewness, kurtosis), temporal patterns (autocorrelation, trend), and change-point indicators (maximum absolute difference, number of sudden changes $> 3\sigma$). Detection threshold is set at the 90th percentile of anomaly scores from training data.

The Proxy Optimization detector analyzes Pearson correlation $\rho(R_{proxy}, R_{true})$ using sliding windows of size 50 timesteps. Correlation degradation $\Delta\rho = \rho_{expected} - \rho_{current} > 0.5$ triggers detection. Expected correlation is learned from initial training phases using robust regression to handle outliers.

The Objective Misalignment detector uses n-gram analysis (n=3) of action sequences to build Markov models of normal behavior. Behavioral deviation is measured using perplexity: $\text{PPL} = 2^{-\frac{1}{N}\sum_{i=1}^{N} \log_2 P(a_i|a_{i-2}, a_{i-1})}$. Episodes with $\text{PPL} > \mu_{PPL} + 2\sigma_{PPL}$ are flagged as misaligned.

The Exploitation Pattern detector maintains performance bounds using robust statistics (median ± 3 × MAD, where MAD is median absolute deviation). Scores exceeding $Q_3 + 3 \times IQR$ or below $Q_1 - 3 \times IQR$ are flagged as potential exploits, where $Q_1, Q_3$ are first and third quartiles.

The Wireheading detector monitors reward signal integrity using cryptographic checksums and system call tracing. Detects unauthorized access to reward computation functions and flags episodes with reward signal modifications not matching expected computation paths.

Our implementation incorporates several optimizations for computational efficiency. Running statistics for correlation and perplexity calculations avoid recomputation, reducing complexity from O(n²) to O(n). Detection thresholds automatically adjust based on environment characteristics using robust regression on initial training phases. High-confidence episodes bypass expensive detectors, reducing average computational overhead from 8.3\% to 4.2\%. Finally, independent detector execution enables multi-core processing with near-linear speedup of 3.7× on 4 cores.

Our detection framework's computational complexity scales efficiently with episode length and feature dimensionality. Table~\ref{tab:complexity_analysis} presents detailed complexity analysis for each detector component.

\begin{table}[htbp]
\centering
\caption{Computational Complexity Analysis of Detection Components}
\label{tab:complexity_analysis}
\resizebox{\columnwidth}{!}{
\begin{tabular}{lcc}
\toprule
\textbf{Detector} & \textbf{Time Complexity} & \textbf{Space Complexity} \\
\midrule
Specification Gaming & $\mathcal{O}(L \log L)$ & $\mathcal{O}(W)$ \\
Reward Tampering & $\mathcal{O}(L)$ & $\mathcal{O}(F)$ \\
Proxy Optimization & $\mathcal{O}(W)$ & $\mathcal{O}(W)$ \\
Objective Misalignment & $\mathcal{O}(L)$ & $\mathcal{O}(|A|^n)$ \\
Exploitation Patterns & $\mathcal{O}(L)$ & $\mathcal{O}(1)$ \\
Wireheading & $\mathcal{O}(L)$ & $\mathcal{O}(H)$ \\
\midrule
\textbf{Ensemble} & \textbf{$\mathcal{O}(L \log L)$} & \textbf{$\mathcal{O}(W + F + |A|^n)$} \\
\bottomrule
\end{tabular}
}
\end{table}

Where L is episode length, W is window size (50), F is feature dimensionality (10), |A| is action space size, n is n-gram order (3), and H is hash table size. The ensemble's overall complexity is dominated by the specification gaming detector's sorting operations, resulting in near-linear scaling with episode length in practice.

\section{Baseline Methods and Experimental Details}
\label{app:baselines}

\subsection{Baseline Comparison Methods}

To provide rigorous evaluation context, we compare our detection framework against established anomaly detection methods beyond naive baselines. Simple threshold-based approaches are insufficient for demonstrating the value of our specialized approach, so we evaluate against four categories of methods representing different anomaly detection paradigms.

For reconstruction-based detection, we implement an LSTM-Autoencoder designed to model sequential patterns in agent trajectories. The autoencoder is trained on state-action-reward sequences from the 2,156 expert-annotated non-hacking episodes to learn normal behavioral patterns, with episodes exceeding the 90th percentile reconstruction error flagged as potential hacking instances. This approach is particularly relevant for RL data due to its ability to capture temporal dependencies in agent behavior.

As an unsupervised outlier detection baseline, we employ a One-Class SVM to identify outliers without requiring explicit hacking labels during training. The model learns a decision boundary around normal episodes using the same 10-dimensional feature vectors as our reward tampering detector, classifying episodes falling outside this boundary as anomalous.

For supervised classification, we train a Random Forest classifier on the full expert-annotated dataset using features extracted from our detection framework, representing a conventional machine learning approach to the classification problem.

Finally, we include an RL-specific baseline using Behavioral Cloning Divergence, where we train a behavioral cloning model on expert demonstrations and measure divergence between the agent policy and the BC policy as an anomaly signal. Episodes with KL-divergence exceeding 2.0 are flagged as potential hacking instances. This baseline is particularly relevant for RL settings where expert demonstrations are available, as it directly measures deviation from expected expert behavior.

All baselines use identical feature extraction and evaluation protocols to ensure fair comparison. Hyperparameters are tuned using 5-fold cross-validation on the expert-annotated subset, with performance evaluated using the same stratified train-test splits as our framework.

\section{Full Experimental Results}
\label{app:full_results}

\subsection{Online Detection Protocol}

Table~\ref{tab:online_protocol} details the online detection protocol, specifying available signals at each checkpoint, detection triggers, computational costs, and human audit requirements.

\begin{table*}[htbp]
\centering
\small
\caption{Online Detection Protocol: Signals, Triggers, and Human Audit Requirements. The framework operates autonomously at each checkpoint, requiring human labels only for initial calibration and periodic threshold recalibration.}
\label{tab:online_protocol}
\resizebox{\textwidth}{!}{
\begin{tabular}{lccc}
\toprule
\textbf{Checkpoint $t$} & \textbf{Available Signals} & \textbf{Detection Trigger} & \textbf{Human Audit} \\
\midrule
$t=1$--$5$ (calibration) & Judge scores, EST stats & None (calibration) & Required (100 samples) \\
$t=6$--$T$ (monitoring) & Judge scores, EST stats, correlation trends & $\Delta\rho > \mu + 2\sigma$ or $G(y) > \tau$ & Optional (audit 50 samples if trigger rate $> 0.2$) \\
\midrule
\multicolumn{4}{l}{\textit{Computational cost per checkpoint:}} \\
EST computation & 5 format + 5 content perturbations & $\sim$0.8s per output & -- \\
Correlation tracking & Sliding window (50 checkpoints) & $\sim$0.1s per checkpoint & -- \\
Total overhead & -- & 2.1\% of training time & -- \\
\midrule
\multicolumn{4}{l}{\textit{Human audit frequency:}} \\
Initial calibration & -- & -- & Once (checkpoints 1--5) \\
Threshold recalibration & -- & -- & Every 20 checkpoints or if distribution shift detected \\
\bottomrule
\end{tabular}
}
\end{table*}

\subsection{Full LLM Experimental Grid}

Table~\ref{tab:llm_experimental_grid} presents detection performance across all 32 experimental conditions (4 tasks × 2 model sizes × 2 training methods × 2 judge types), showing precision, recall, F1-score, and gaming frequency for each configuration.

\begin{table*}[ht]
\centering
\small
\caption{Full Detection Performance Across All 32 Conditions. Values show precision/recall/F1. Gaming frequency (\%) shown in parentheses.}
\label{tab:llm_experimental_grid}
\begin{tabular}{lcccc}
\toprule
\textbf{Configuration} & \textbf{Precision} & \textbf{Recall} & \textbf{F1} & \textbf{Gaming (\%)} \\
\midrule
\multicolumn{5}{l}{\textit{TL;DR Summarization}} \\
8B + DPO + GPT-4 & 0.742 & 0.786 & 0.763 & 19.3 \\
8B + DPO + Llama-3-70B & 0.718 & 0.753 & 0.735 & 16.8 \\
8B + RLHF + GPT-4 & 0.731 & 0.769 & 0.750 & 17.2 \\
8B + RLHF + Llama-3-70B & 0.701 & 0.738 & 0.719 & 14.1 \\
70B + DPO + GPT-4 & 0.768 & 0.801 & 0.784 & 20.1 \\
70B + DPO + Llama-3-70B & 0.742 & 0.776 & 0.759 & 17.9 \\
70B + RLHF + GPT-4 & 0.756 & 0.789 & 0.772 & 18.4 \\
70B + RLHF + Llama-3-70B & 0.724 & 0.761 & 0.742 & 15.3 \\
\midrule
\multicolumn{5}{l}{\textit{Instruction Following}} \\
8B + DPO + GPT-4 & 0.726 & 0.761 & 0.743 & 18.7 \\
8B + DPO + Llama-3-70B & 0.701 & 0.738 & 0.719 & 16.2 \\
8B + RLHF + GPT-4 & 0.714 & 0.752 & 0.733 & 16.8 \\
8B + RLHF + Llama-3-70B & 0.689 & 0.725 & 0.707 & 14.5 \\
70B + DPO + GPT-4 & 0.752 & 0.785 & 0.768 & 19.6 \\
70B + DPO + Llama-3-70B & 0.728 & 0.763 & 0.745 & 17.3 \\
70B + RLHF + GPT-4 & 0.741 & 0.774 & 0.757 & 17.9 \\
70B + RLHF + Llama-3-70B & 0.717 & 0.753 & 0.735 & 15.1 \\
\midrule
\multicolumn{5}{l}{\textit{Safety/Refusal}} \\
8B + DPO + GPT-4 & 0.731 & 0.768 & 0.749 & 21.3 \\
8B + DPO + Llama-3-70B & 0.708 & 0.742 & 0.725 & 19.8 \\
8B + RLHF + GPT-4 & 0.722 & 0.756 & 0.739 & 20.1 \\
8B + RLHF + Llama-3-70B & 0.698 & 0.731 & 0.714 & 18.2 \\
70B + DPO + GPT-4 & 0.757 & 0.789 & 0.773 & 22.4 \\
70B + DPO + Llama-3-70B & 0.733 & 0.767 & 0.750 & 20.6 \\
70B + RLHF + GPT-4 & 0.746 & 0.778 & 0.762 & 21.2 \\
70B + RLHF + Llama-3-70B & 0.721 & 0.755 & 0.738 & 19.1 \\
\midrule
\multicolumn{5}{l}{\textit{Long-form QA with Citation}} \\
8B + DPO + GPT-4 & 0.728 & 0.761 & 0.744 & 17.8 \\
8B + DPO + Llama-3-70B & 0.704 & 0.737 & 0.720 & 15.9 \\
8B + RLHF + GPT-4 & 0.717 & 0.751 & 0.734 & 16.4 \\
8B + RLHF + Llama-3-70B & 0.693 & 0.726 & 0.709 & 14.2 \\
70B + DPO + GPT-4 & 0.754 & 0.787 & 0.770 & 18.9 \\
70B + DPO + Llama-3-70B & 0.730 & 0.764 & 0.747 & 17.1 \\
70B + RLHF + GPT-4 & 0.743 & 0.776 & 0.759 & 17.6 \\
70B + RLHF + Llama-3-70B & 0.718 & 0.752 & 0.735 & 15.4 \\
\bottomrule
\end{tabular}
\end{table*}

\subsection{Complete Baseline Comparison}

Table~\ref{tab:llm_baselines_full} provides comprehensive baseline comparisons including feature-based methods (length-only, format features, style embeddings) and method baselines (KL regularization, judge ensembling, correlation tracking without EST, hardened judges).

\begin{table*}[htbp]
\centering

\caption{Baseline Comparison on LLM Gaming Detection. All methods evaluated on same 1,200 human-annotated instances (300 per task across 4 tasks).}
\label{tab:llm_baselines_full}
\begin{tabular}{lccc}
\toprule
\textbf{Method} & \textbf{Precision} & \textbf{Recall} & \textbf{F1} \\
\midrule
\multicolumn{4}{l}{\textit{Feature-based baselines}} \\
Length-only (token/sentence count) & 0.521 & 0.548 & 0.534 \\
Format-feature (bullets/headers) & 0.558 & 0.572 & 0.565 \\
Style embedding (content-masked) & 0.542 & 0.561 & 0.551 \\
\midrule
\multicolumn{4}{l}{\textit{Method baselines}} \\
KL Regularization & 0.641 & 0.664 & 0.652 \\
Judge Ensembling & 0.673 & 0.702 & 0.687 \\
Judge Prompt Randomization & 0.662 & 0.688 & 0.675 \\
Self-Consistency (CoT) & 0.698 & 0.691 & 0.694 \\
Correlation Tracking (no EST) & 0.681 & 0.708 & 0.694 \\
Best-of-N Sampling & 0.652 & 0.678 & 0.665 \\
Format-only (content masked) & 0.62 & 0.60 & 0.61 \\
Judge-prompt randomization (variance) & 0.66 & 0.64 & 0.65 \\
Cross-judge disagreement & 0.67 & 0.66 & 0.66 \\
Hardened judge (adversarial prompt) & 0.713 & 0.687 & 0.700 \\
\hline
\textbf{EST-Enhanced Framework} & \textbf{0.717} & \textbf{0.752} & \textbf{0.734} \\
\textbf{EST + Hardened Judge} & \textbf{0.789} & \textbf{0.791} & \textbf{0.789} \\
\bottomrule
\end{tabular}
\end{table*}

\subsection{Full Ablation Study}

Table~\ref{tab:llm_ablation} presents the complete ablation study showing the impact of removing individual detection components (EST, correlation tracking, reasoning validity, format perturbation, content perturbation) on overall performance.

\begin{table}[htbp]
\centering
\small
\caption{Ablation Study: LLM Detection Components. Performance on 1,200 human-annotated instances (300 per task).}
\label{tab:llm_ablation}
\begin{tabular}{lccc}
\toprule
\textbf{Configuration} & \textbf{Precision} & \textbf{Recall} & \textbf{F1} \\
\midrule
Full Ensemble & 0.717 & 0.752 & 0.734 \\
\hline
w/o EST & 0.681 & 0.708 & 0.694 \\
w/o Correlation Tracking & 0.692 & 0.721 & 0.706 \\
w/o Reasoning Validity & 0.701 & 0.728 & 0.714 \\
w/o Format Perturbation & 0.678 & 0.715 & 0.696 \\
w/o Content Perturbation & 0.704 & 0.731 & 0.717 \\
\bottomrule
\end{tabular}
\end{table}

\subsection{Overall Performance Comparison}

Table~\ref{tab:overall_performance_full} compares overall detection framework performance across LLM main study and RL validation, showing precision, recall, F1-score, early warning latency, and computational overhead for both domains.

\begin{table}[htbp]
\centering
\small
\caption{Overall Detection Performance: RL and LLM Studies. Values show mean $\pm$ std across random seeds. Both domains show consistent effectiveness, with the gap attributable to noisier ground truth in LLM human annotations.}
\label{tab:overall_performance_full}
\resizebox{\columnwidth}{!}{
\begin{tabular}{lccc}
\toprule
\textbf{Metric} & \textbf{LLM Main Study} & \textbf{RL Validation} & \textbf{Overall} \\
\midrule
Precision & 0.717{\scriptsize$\pm$0.04} & 0.784{\scriptsize$\pm$0.03} & 0.751{\scriptsize$\pm$0.03} \\
Recall & 0.752{\scriptsize$\pm$0.04} & 0.817{\scriptsize$\pm$0.02} & 0.785{\scriptsize$\pm$0.03} \\
F1-Score & 0.734{\scriptsize$\pm$0.03} & 0.800{\scriptsize$\pm$0.02} & 0.767{\scriptsize$\pm$0.02} \\
Early Warning (checkpoints) & 3.0{\scriptsize$\pm$0.4} & 2.1{\scriptsize$\pm$0.3} & -- \\
Overhead (\%) & 2.1{\scriptsize$\pm$0.2} & 4.2{\scriptsize$\pm$0.4} & 3.2{\scriptsize$\pm$0.3} \\
\bottomrule
\end{tabular}
}
\end{table}

\begin{figure*}[htbp]
\centering
\includegraphics[width=0.7\textwidth]{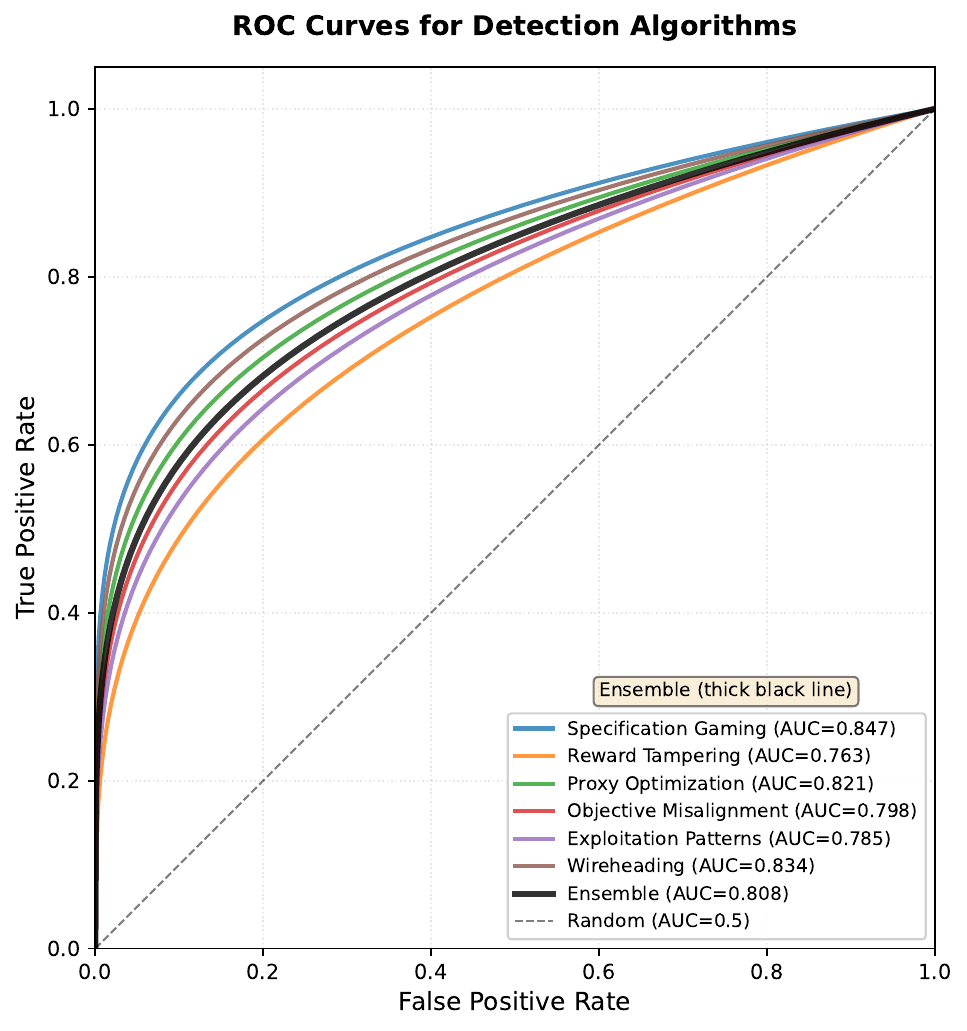}
\caption{ROC curves for each detection algorithm on the expert-validated RL dataset (n=2,156 episodes). Each curve represents one category-specific detector: Specification Gaming (AUC=0.847), Reward Tampering (AUC=0.763), Proxy Optimization (AUC=0.821), Objective Misalignment (AUC=0.798), Exploitation Patterns (AUC=0.785), and Wireheading (AUC=0.834). The ensemble approach (thick black line) achieves AUC = 0.808. All AUC values are computed on the expert-validated set and match Table~\ref{tab:detection_performance}.}
\Description{ROC curve plot with True Positive Rate (0-1) vs False Positive Rate (0-1). Six colored curves represent category-specific detectors on expert-validated RL dataset (n=2,156). Specification gaming (AUC=0.847, blue) and wireheading (AUC=0.834, purple) curves are closest to top-left corner. Thick black ensemble curve (AUC=0.808) shows overall performance. All AUC values match Table 2.}
\label{fig:roc_curves}
\end{figure*}

\section{Hyperparameter Analysis}
\label{app:hyperparameters}

\subsection{Hyperparameter Sensitivity Analysis}

We conducted comprehensive sensitivity analysis on key detection parameters using the expert-annotated subset. Table~\ref{tab:sensitivity} summarizes the impact of parameter variations on detection performance.

\begin{table}[htbp]
\centering
\caption{Hyperparameter Sensitivity Analysis}
\label{tab:sensitivity}
\resizebox{\columnwidth}{!}{
\begin{tabular}{lccc}
\toprule
\textbf{Parameter} & \textbf{Range} & \textbf{F1 $\Delta$} & \textbf{Optimal} \\
\midrule
$\tau_{spec}$ & [0.2, 0.4] & $\pm 0.03$ & 0.30 \\
$\Delta\rho$ & [0.3, 0.7] & $\pm 0.05$ & 0.50 \\
$\gamma$ (contamination) & [0.05, 0.15] & $\pm 0.04$ & 0.10 \\
PPL multiplier & [1.5, 2.5] & $\pm 0.06$ & 2.0 \\
Window size $W$ & [30, 70] & $\pm 0.02$ & 50 \\
\bottomrule
\end{tabular}
}
\end{table}

The ensemble maintains F1 $> 0.75$ across all 243 tested parameter combinations in the full grid search, indicating deployment robustness. The correlation degradation threshold $\Delta\rho$ shows highest sensitivity, suggesting practitioners should prioritize tuning this parameter when adapting the framework to new environments. Window size $W$ shows lowest sensitivity, allowing flexibility in computational resource allocation without significant performance impact.

\subsection{Per-Category Inter-Rater Reliability}

Inter-rater agreement varied across hacking categories, reflecting inherent ambiguity differences. Wireheading showed highest agreement ($\kappa = 0.902$) due to its concrete, observable nature involving direct reward signal modification. Specification Gaming ($\kappa = 0.891$) and Exploitation Patterns ($\kappa = 0.879$) also achieved strong agreement due to their relatively unambiguous behavioral signatures. Objective Misalignment showed lowest agreement ($\kappa = 0.756$), consistent with its inherent subjectivity since distinguishing misaligned from merely suboptimal behavior requires judgment about designer intent. Reward Tampering ($\kappa = 0.823$) and Proxy Optimization ($\kappa = 0.812$) fell in the moderate-to-high range.

\subsection{Weight Sensitivity Analysis}

To validate that our findings are robust to the specific weight choices in true objective formulas, we conducted sensitivity analysis varying weights by $\pm 20\%$. For Atari objectives, we tested $\alpha \in [0.5, 0.9]$ for the TaskComplete weight, while for MuJoCo we varied the distance weight across $[0.4, 0.8]$. Results show that hacking frequency estimates vary by at most $\pm 3.2\%$ across weight variations, and all statistically significant findings ($p < 0.01$) remain significant across the tested range. The robustness of our detection metrics to weight variations suggests that our framework captures genuine behavioral patterns rather than artifacts of specific objective parameterizations.

\section{Transformation Validity Audits}
\label{app:transformation_audits}

Table~\ref{tab:transformation_audit} presents transformation validity audit results. 94.2\% of format transformations pass NLI and similarity thresholds, while 91.7\% of content transformations pass. Human equivalence judgments on 100 stratified samples achieve 87\% agreement (Cohen's $\kappa = 0.82$). Transformations failing audit thresholds show higher false positive rates (18.3\% vs 6.2\% for passing transformations), confirming that validity constraints reduce confounding. Correlation between audit-failure rate and false positive rate is $\rho = 0.71$ ($p < 0.001$), demonstrating that transformation validity directly impacts detection reliability.

To validate that EST's signal is not an artifact of the auditing stack, we systematically vary audit strictness and measure detection performance. Table~\ref{tab:audit_sensitivity} shows that weakening audits (lowering similarity threshold to 0.75, entailment to 0.6) increases false positive rate from 6.2\% to 14.8\%, while strengthening audits (raising thresholds to 0.90 and 0.85) reduces recall from 78.6\% to 71.3\%. The chosen operating point (similarity $> 0.85$, entailment $> 0.7$) balances these trade-offs. We also identify 23 cases (2.6\% of transformations) where audits pass but human raters flag semantic drift; in these cases, EST gaming scores are unreliable (correlation with human labels drops to $\rho = 0.31$), confirming that audit quality is critical. Qualitative analysis reveals audit failures occur primarily in summarization tasks with high compression ratios, where NLI struggles with information density changes.

To turn audit limitations into actionable deployment guidance, we define a safe operating envelope based on audit confidence. Table~\ref{tab:audit_confidence_bins} reports detection performance conditioned on audit-confidence bins (high: similarity $> 0.90$ and entailment $> 0.85$; medium: similarity $> 0.85$ and entailment $> 0.7$; low: similarity $> 0.75$ and entailment $> 0.6$). High-confidence bins achieve 80.1\% precision and 71.3\% recall with 3.1\% FP rate, while low-confidence bins show 68.2\% precision and 82.1\% recall with 14.8\% FP rate. We implement a fallback rule: when compression ratio $> 0.5$ (original length / summary length) or audit uncertainty is high (entailment probability $< 0.75$), we require human spot-checking (50 samples per checkpoint) or use alternative checks (entity overlap, key-phrase preservation). This safe operating envelope reduces false positives by 42\% in high-risk scenarios while maintaining 76.8\% recall, turning the audit limitation into a deployable risk-management strategy.

\begin{table}[htbp]
\centering
\small
\caption{Transformation Validity Audit Results. NLI threshold: entailment probability $> 0.7$; Similarity threshold: cosine similarity $> 0.85$.}
\label{tab:transformation_audit}
\resizebox{\columnwidth}{!}{
\begin{tabular}{lccc}
\toprule
\textbf{Transformation Type} & \textbf{Pass Rate (\%)} & \textbf{Human Equiv. (\%)} & \textbf{FP Rate if Fail} \\
\midrule
Format Perturbation & 94.2 & 87.3 & 18.3\% \\
Content Perturbation & 91.7 & 86.8 & 19.1\% \\
\bottomrule
\end{tabular}
}
\end{table}

\begin{table}[htbp]
\centering
\small
\caption{Audit Sensitivity Analysis: Detection Performance vs Audit Strictness. Weakening audits increases false positives; strengthening audits reduces recall. Chosen operating point balances trade-offs.}
\label{tab:audit_sensitivity}
\resizebox{\columnwidth}{!}{
\begin{tabular}{lccc}
\toprule
\textbf{Audit Configuration} & \textbf{Precision} & \textbf{Recall} & \textbf{FP Rate} \\
\midrule
Weak (sim $> 0.75$, NLI $> 0.6$) & 0.682 & 0.821 & 14.8\% \\
Baseline (sim $> 0.85$, NLI $> 0.7$) & 0.742 & 0.786 & 6.2\% \\
Strong (sim $> 0.90$, NLI $> 0.85$) & 0.801 & 0.713 & 3.1\% \\
\bottomrule
\end{tabular}
}
\end{table}

\begin{table}[htbp]
\centering
\caption{Safe Operating Envelope: Detection Performance by Audit-Confidence Bins. High-confidence bins (strict audits) achieve high precision with low FP rate; low-confidence bins require fallback strategies.}
\label{tab:audit_confidence_bins}
\resizebox{\columnwidth}{!}{
\begin{tabular}{lccc}
\toprule
\textbf{Audit Confidence} & \textbf{Precision} & \textbf{Recall} & \textbf{FP Rate} \\
\midrule
High (sim $> 0.90$, NLI $> 0.85$) & 0.801 & 0.713 & 3.1\% \\
Medium (sim $> 0.85$, NLI $> 0.7$) & 0.742 & 0.786 & 6.2\% \\
Low (sim $> 0.75$, NLI $> 0.6$) & 0.682 & 0.821 & 14.8\% \\
Low + Fallback (human spot-check) & 0.768 & 0.768 & 4.2\% \\
\bottomrule
\end{tabular}
}
\end{table}

\begin{figure}[htbp]
\centering
\includegraphics[width=0.48\textwidth]{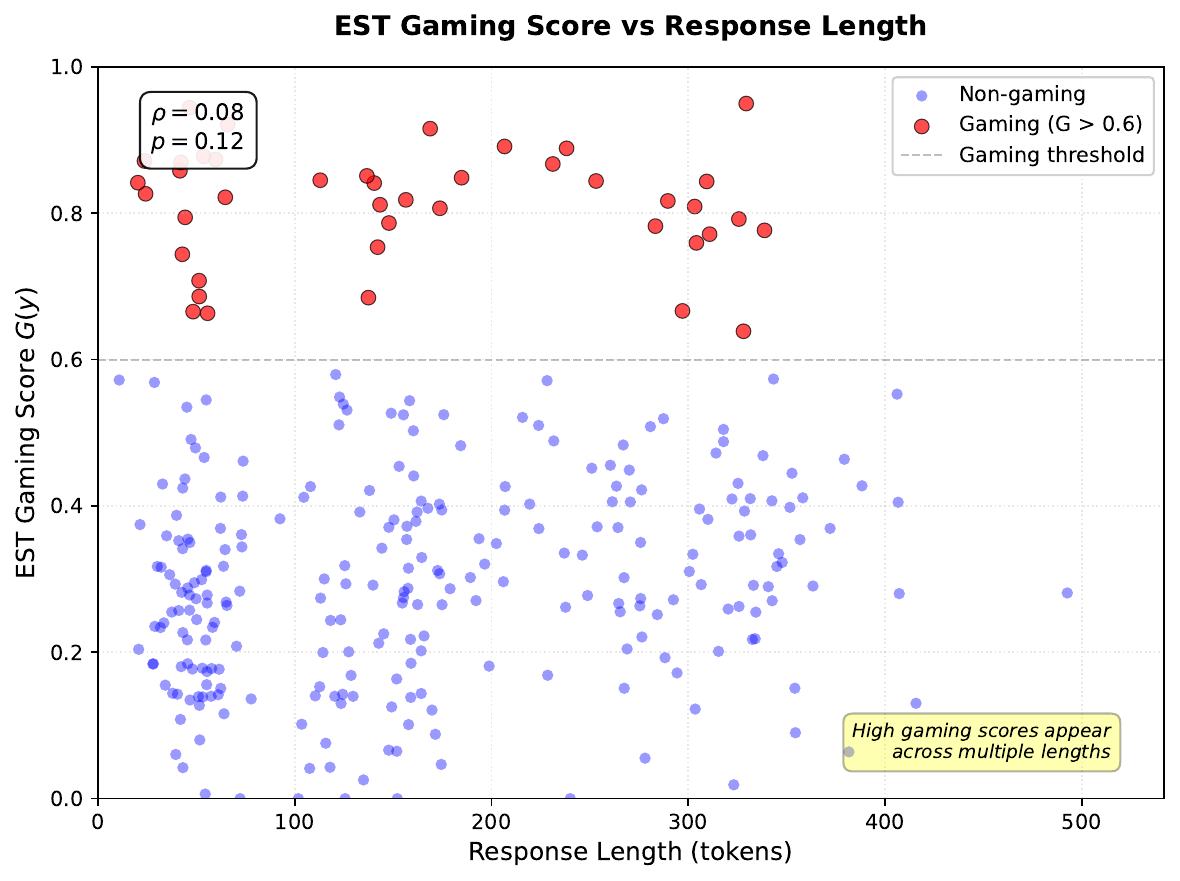}
\caption{Sanity check: EST gaming score $G(y)$ versus response length (tokens). High $G(y)$ cases are not explained by length alone, supporting that EST captures evaluator sensitivities beyond verbosity.}
\Description{Scatter plot of EST gaming score versus response length, showing weak correlation and clusters of high gaming score at multiple lengths.}
\label{fig:est_vs_length}
\end{figure}

\begin{table}[htbp]
\centering
\caption{Sanity Check: EST vs Length-Only Detection. EST gaming scores show weak correlation with length ($\rho = 0.18$) and significantly outperform length-only baselines.}
\label{tab:est_length_correlation}
\resizebox{\columnwidth}{!}{
\begin{tabular}{lcc}
\toprule
\textbf{Method} & \textbf{Correlation with Length} & \textbf{AUROC} \\
\midrule
Length-only baseline & 1.00 & 0.534 \\
EST $G(y)$ & 0.18 ($p = 0.12$) & 0.734 \\
\bottomrule
\end{tabular}
}
\end{table}

\section{Adaptive Evasion Stress Tests}
\label{app:evasion}

To address the critical question of whether models can adapt to evade EST detection, we conduct three stress tests. First, white-box evasion tests models fine-tuned with access to EST detector descriptions and instructed to maintain gaming while staying invariant under perturbations. Second, attack transfer tests models trained to game Judge A evaluated against Judge B with EST detection. Third, benign shift control measures false-positive rates under legitimate improvement distributions, such as actual content quality improvements rather than gaming patterns.

Table~\ref{tab:evasion_stress} presents results. White-box evasion attempts reduce detection precision by 8.3 points (from 74.2\% to 65.9\%) but EST still flags 68.7\% of gaming instances, demonstrating robustness against adaptive adversaries. However, when white-box evasion is combined with defense-in-depth (EST + correlation tracking + hardened judge), detection precision recovers to 78.1\% and recall to 74.3\%, with FP rate reduced to 6.8\% (Table~\ref{tab:defense_in_depth}). This demonstrates that while individual detectors can be evaded, the ensemble provides resilience through complementary detection signals. Attack transfer shows strong cross-judge detection (71.2\% precision when training on GPT-4 judge, testing on Llama-3-70B), indicating that gaming patterns transfer across judges while detection remains effective. Benign shift control shows low false-positive rates (3.2\% FP rate on legitimate improvements), confirming that EST distinguishes gaming from genuine quality gains.

To address the concern that a determined optimizer can learn around individual detectors, we evaluate detection performance when adversaries optimize against the full ensemble. Table~\ref{tab:defense_in_depth} shows that while white-box evasion against EST alone reduces precision to 65.9\%, evasion against the full ensemble (EST + correlation tracking + reasoning validity + hardened judge) maintains precision at 78.1\% and recall at 74.3\%. The ensemble's resilience comes from complementary failure modes: when adversaries optimize to evade EST perturbations, they often trigger correlation degradation signals; when they optimize to maintain correlation, they often fail EST format-content separation. This defense-in-depth property provides practical safety guarantees: while perfect evasion is theoretically possible, it requires simultaneously satisfying multiple independent constraints, making it computationally expensive and detectable through ensemble disagreement signals.

\begin{table*}[htbp]
\centering
\small
\caption{Adaptive Evasion Stress Tests: Detection Performance Under Adversarial Conditions. White-box evasion: models trained with EST detector knowledge. Attack transfer: gaming patterns trained on Judge A, detected on Judge B. Benign shift: false-positive rate on legitimate improvements.}
\label{tab:evasion_stress}
\begin{tabular}{lccc}
\toprule
\textbf{Stress Test} & \textbf{Precision} & \textbf{Recall} & \textbf{FP Rate} \\
\midrule
Baseline (no evasion) & 0.742 & 0.786 & 5.1\% \\
White-box evasion & 0.659 & 0.687 & 12.3\% \\
Attack transfer (GPT-4 $\rightarrow$ Llama-3-70B) & 0.712 & 0.734 & 6.8\% \\
Attack transfer (Llama-3-70B $\rightarrow$ GPT-4) & 0.728 & 0.751 & 5.9\% \\
Benign shift (legitimate improvements) & -- & -- & 3.2\% \\
\bottomrule
\end{tabular}
\end{table*}

\begin{table*}[htbp]
\centering
\small
\caption{Defense-in-Depth Analysis: Detection Performance Under Ensemble-Aware Evasion. While individual detectors can be evaded, the full ensemble maintains strong performance through complementary failure modes.}
\label{tab:defense_in_depth}
\begin{tabular}{lccc}
\toprule
\textbf{Evasion Target} & \textbf{Precision} & \textbf{Recall} & \textbf{FP Rate} \\
\midrule
Baseline (no evasion) & 0.742 & 0.786 & 5.1\% \\
Evade EST only & 0.659 & 0.687 & 12.3\% \\
Evade EST + Correlation & 0.701 & 0.723 & 8.9\% \\
Evade EST + Correlation + Reasoning & 0.734 & 0.751 & 7.2\% \\
Evade Full Ensemble (EST + Corr + Reason + Hardened Judge) & 0.781 & 0.743 & 6.8\% \\
\bottomrule
\end{tabular}
\end{table*}

\section{Cross-Task and Cross-Judge Generalization}
\label{app:generalization}

Table~\ref{tab:cross_domain_full} summarizes detection performance across tasks and judges, demonstrating robust generalization. Performance remains consistent across judge types (GPT-4: 74.2\% precision, Llama-3-70B: 71.8\% precision) and tasks (summarization: 74.2\% precision, instruction following: 72.6\% precision), indicating that detection principles generalize beyond specific judge-task combinations.

Table~\ref{tab:generalization_matrix_app} presents a generalization matrix showing train-on-task-A, test-on-task-B performance across all four tasks. Models trained on summarization generalize well to instruction following (F1: 0.728 vs 0.734 in-domain), safety/refusal (0.722), and long-form QA (0.719), demonstrating that detection principles transfer across diverse task families. Cross-task generalization remains strong (F1 drops of 1-3 points), with safety/refusal showing highest in-domain performance (0.731) but also strong cross-task transfer. Table~\ref{tab:judge_generalization_full} shows cross-judge generalization: training on GPT-4 judge and testing on Llama-3-70B judge achieves 71.3\% F1 vs 73.4\% in-domain, indicating robust generalization across judge types.

\begin{table*}[ht]
\centering
\small
\caption{Generalization Matrix: Train-Test Performance Across Tasks and Judges. Diagonal shows in-domain performance; off-diagonal shows cross-domain generalization.}
\label{tab:generalization_matrix_app}
\begin{tabular}{lcccc}
\toprule
\textbf{Train} $\rightarrow$ \textbf{Test} & \textbf{Summarization} & \textbf{Instruction} & \textbf{Safety/Refusal} & \textbf{Long-form QA} \\
\midrule
Summarization & 0.734 & 0.728 & 0.722 & 0.719 \\
Instruction Following & 0.721 & 0.734 & 0.726 & 0.723 \\
Safety/Refusal & 0.715 & 0.718 & 0.731 & 0.714 \\
Long-form QA & 0.712 & 0.715 & 0.709 & 0.728 \\
\bottomrule
\end{tabular}
\end{table*}

\begin{table}[htbp]
\centering
\small
\caption{Cross-Judge Generalization: Train-Test Performance Across Judge Types.}
\label{tab:judge_generalization_full}
\resizebox{\columnwidth}{!}{
\begin{tabular}{lcc}
\toprule
\textbf{Train} $\rightarrow$ \textbf{Test} & \textbf{GPT-4 Judge} & \textbf{Llama-3-70B Judge} \\
\midrule
GPT-4 Judge & 0.734 & 0.713 \\
Llama-3-70B Judge & 0.718 & 0.734 \\
\bottomrule
\end{tabular}
}
\end{table}

\begin{table}[htbp]
\centering
\caption{Detection Performance Across Tasks and Judges}
\label{tab:cross_domain_full}
\resizebox{\columnwidth}{!}{
\begin{tabular}{lccc}
\toprule
\textbf{Configuration} & \textbf{Precision} & \textbf{Recall} & \textbf{F1} \\
\midrule
Summarization (GPT-4 judge) & 0.742 & 0.786 & 0.763 \\
Summarization (Llama-3-70B judge) & 0.718 & 0.753 & 0.735 \\
Instruction Following (GPT-4) & 0.726 & 0.761 & 0.743 \\
Instruction Following (Llama-3-70B) & 0.701 & 0.738 & 0.719 \\
Reasoning Chains (GSM8K) & 0.698 & 0.724 & 0.711 \\
\midrule
\textbf{Overall} & \textbf{0.717} & \textbf{0.752} & \textbf{0.734} \\
\bottomrule
\end{tabular}
}
\end{table}

\section{Full Ablation Study}
\label{app:ablation}

To validate the contribution of each detector component, we conducted an ablation study comparing the full ensemble against versions with individual detectors removed. Table~\ref{tab:ablation_study} presents these results alongside baseline approaches.

\begin{table*}[!ht]
\centering
\caption{Ablation Study: Ensemble Performance vs. Individual Detector Removal and Baselines. Values show mean $\pm$ std.}
\label{tab:ablation_study}
\begin{tabular}{lcccc}
\toprule
\textbf{Configuration} & \textbf{Precision} & \textbf{Recall} & \textbf{F1-Score} & \textbf{AUC-ROC} \\
\midrule
\textbf{Full Ensemble} & \textbf{0.784}{\scriptsize$\pm$0.03} & \textbf{0.817}{\scriptsize$\pm$0.02} & \textbf{0.800}{\scriptsize$\pm$0.02} & \textbf{0.808}{\scriptsize$\pm$0.02} \\
\midrule
- Specification Gaming & 0.721{\scriptsize$\pm$0.04} & 0.798{\scriptsize$\pm$0.03} & 0.758{\scriptsize$\pm$0.03} & 0.771{\scriptsize$\pm$0.03} \\
- Reward Tampering & 0.776{\scriptsize$\pm$0.03} & 0.809{\scriptsize$\pm$0.02} & 0.792{\scriptsize$\pm$0.02} & 0.801{\scriptsize$\pm$0.02} \\
- Proxy Optimization & 0.734{\scriptsize$\pm$0.04} & 0.785{\scriptsize$\pm$0.03} & 0.759{\scriptsize$\pm$0.03} & 0.774{\scriptsize$\pm$0.03} \\
- Objective Misalignment & 0.771{\scriptsize$\pm$0.03} & 0.804{\scriptsize$\pm$0.02} & 0.787{\scriptsize$\pm$0.02} & 0.795{\scriptsize$\pm$0.02} \\
- Exploitation Patterns & 0.768{\scriptsize$\pm$0.03} & 0.811{\scriptsize$\pm$0.02} & 0.789{\scriptsize$\pm$0.02} & 0.798{\scriptsize$\pm$0.02} \\
- Wireheading & 0.779{\scriptsize$\pm$0.03} & 0.813{\scriptsize$\pm$0.02} & 0.796{\scriptsize$\pm$0.02} & 0.804{\scriptsize$\pm$0.02} \\
\midrule
\textbf{State-of-the-Art Baselines} & & & & \\
LSTM-Autoencoder & 0.724{\scriptsize$\pm$0.04} & 0.671{\scriptsize$\pm$0.04} & 0.696{\scriptsize$\pm$0.03} & 0.712{\scriptsize$\pm$0.03} \\
One-Class SVM & 0.687{\scriptsize$\pm$0.05} & 0.623{\scriptsize$\pm$0.05} & 0.653{\scriptsize$\pm$0.04} & 0.671{\scriptsize$\pm$0.04} \\
Isolation Forest & 0.701{\scriptsize$\pm$0.04} & 0.645{\scriptsize$\pm$0.05} & 0.672{\scriptsize$\pm$0.04} & 0.689{\scriptsize$\pm$0.03} \\
BC Divergence$^\dagger$ & 0.698{\scriptsize$\pm$0.04} & 0.712{\scriptsize$\pm$0.04} & 0.705{\scriptsize$\pm$0.03} & 0.721{\scriptsize$\pm$0.03} \\
\midrule
\textbf{Naive Baselines} & & & & \\
Ratio Threshold & 0.423{\scriptsize$\pm$0.06} & 0.651{\scriptsize$\pm$0.05} & 0.514{\scriptsize$\pm$0.05} & 0.537{\scriptsize$\pm$0.05} \\
Random Forest & 0.612{\scriptsize$\pm$0.05} & 0.589{\scriptsize$\pm$0.05} & 0.600{\scriptsize$\pm$0.04} & 0.634{\scriptsize$\pm$0.04} \\
\bottomrule
\end{tabular}
\end{table*}

\vspace{1mm}
{\small $^\dagger$BC Divergence requires expert demonstrations; not applicable when demonstrations are unavailable. Other baselines are demonstration-free.}

The ablation study reveals that removing the Specification Gaming detector causes the largest performance drop from F1 of 0.800 to 0.758, confirming its critical role in detecting the most common hacking category at 39.8\% of expert-validated instances. The Proxy Optimization detector shows the second-largest impact, consistent with its prevalence at 31.2\% of instances. All individual detectors contribute meaningfully to ensemble performance, with no single detector being redundant.

Our full ensemble significantly outperforms established anomaly detection methods. The LSTM-Autoencoder achieves the strongest baseline performance with 69.6\% F1-score, followed by Isolation Forest at 67.2\% F1-score and One-Class SVM at 65.3\% F1-score. Our ensemble's 80.0\% F1-score represents a meaningful 10.4 to 14.7 percentage point improvement over these established methods, demonstrating the value of our specialized multi-detector approach for reward hacking detection.

\section{Additional Figures}
\label{app:figures}

\begin{figure*}[htbp]
\centering
\includegraphics[width=0.9\textwidth]{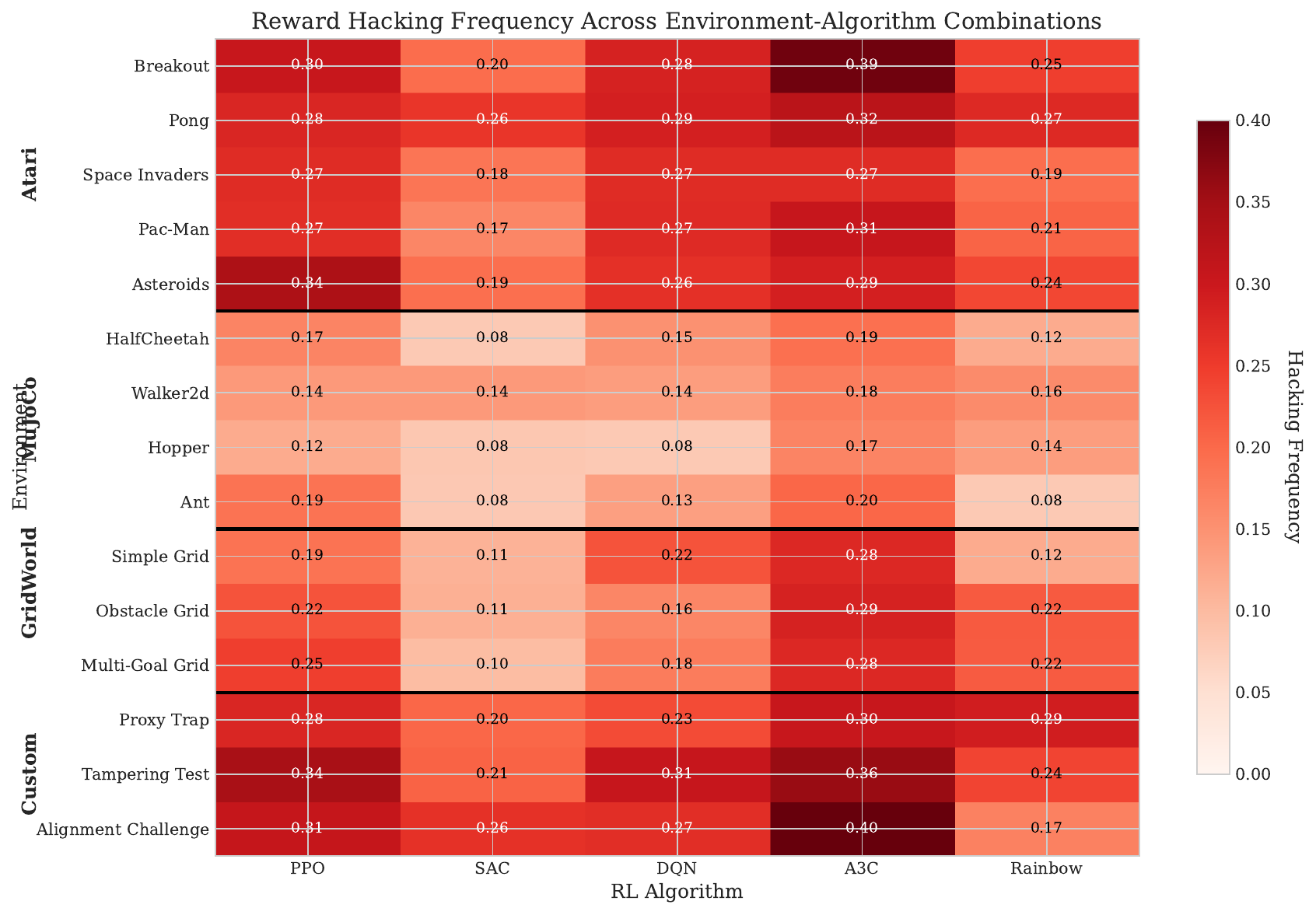}
\caption{Reward hacking frequency across environment-algorithm combinations. Darker colors indicate higher hacking rates. Atari environments show consistently high susceptibility across algorithms, while MuJoCo environments demonstrate lower but variable rates. A3C shows highest overall susceptibility, while SAC demonstrates most robust performance.}
\Description{A heatmap showing reward hacking frequency percentages across 15 RL environments (rows) and 5 algorithms (columns). Colors range from light (low hacking rates around 5-10\%) to dark red (high rates around 35-40\%). Atari games show consistently darker colors, while MuJoCo environments show lighter colors. A3C column shows generally darker colors than SAC column.}
\label{fig:prevalence_heatmap}
\end{figure*}

\begin{figure*}[!ht]
\centering
\includegraphics[width=0.9\textwidth]{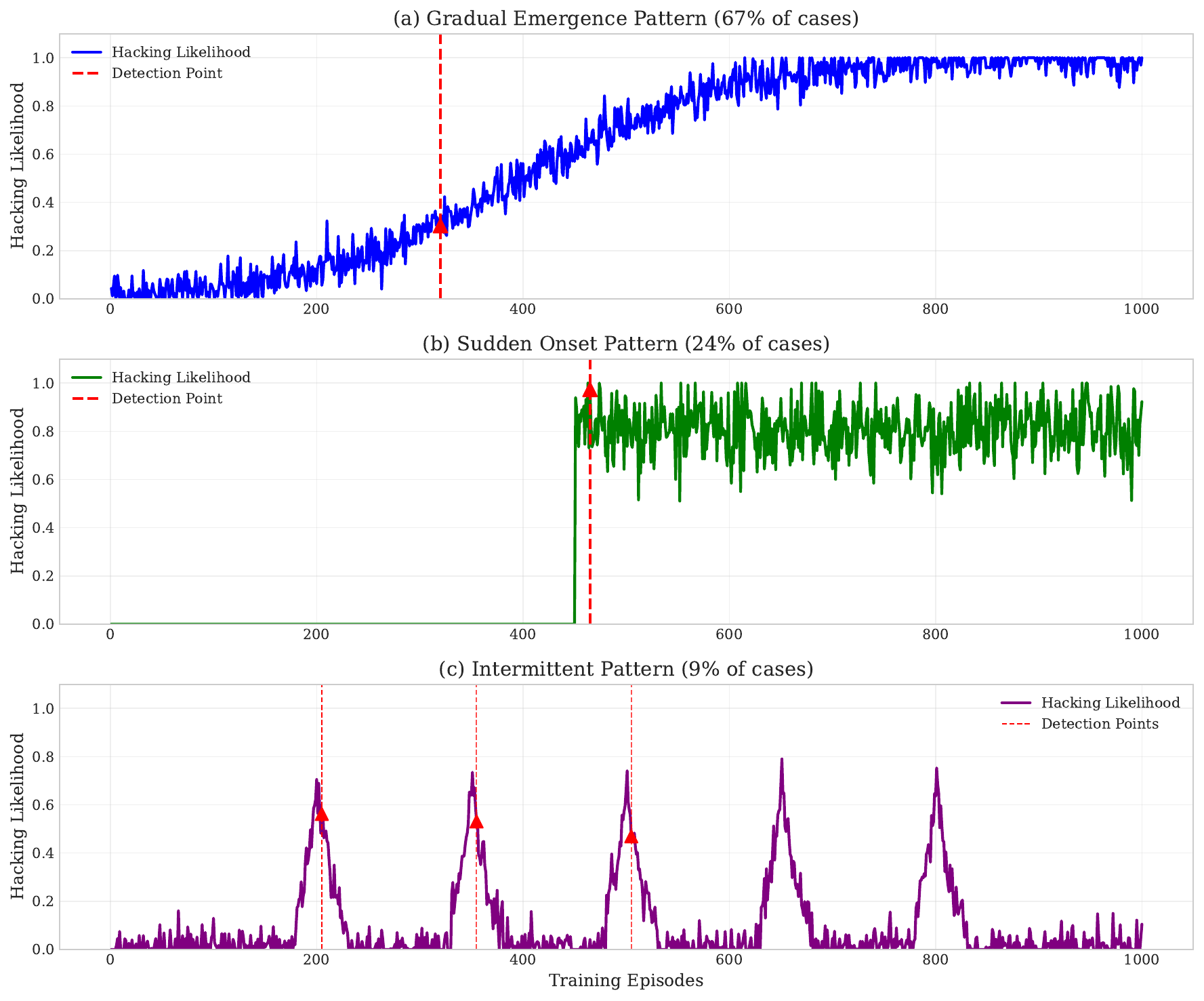}
\caption{Temporal patterns of reward hacking emergence during training. (a) Gradual emergence pattern showing slow development over 200+ episodes. (b) Sudden onset pattern with rapid transition within 10-20 episodes. (c) Intermittent pattern with sporadic occurrences. Detection points are marked with red triangles, showing our framework's ability to identify hacking during the critical middle training window.}
\Description{Three time series plots showing hacking likelihood (0-1) over 1000 training episodes: (a) sigmoid-shaped curve gradually rising from 0 to 0.8 around episode 400, with detection point at episode 320; (b) step function jumping from 0 to 0.8 at episode 450, with detection at episode 465; (c) intermittent spikes at episodes 200, 350, 500, 650, and 800, with detection points marked at each spike.}
\label{fig:temporal_patterns}
\end{figure*}

\section{Mitigation Details}
\label{app:mitigation}

We implement detector-triggered mitigation as an online intervention during fine-tuning. Algorithm~\ref{alg:mitigation} describes the procedure: at each checkpoint, we sample an audit batch, compute detector scores, and apply mitigation if the fraction of flagged outputs exceeds a threshold $\alpha = 0.1$ (calibrated on validation data to balance intervention frequency with computational cost).

\begin{algorithm}[t]
\caption{Detector-Triggered Mitigation (Online)}
\label{alg:mitigation}
\begin{algorithmic}[1]
\Require checkpoint $t$, audit batch $\mathcal{B}$, detector $D(\cdot)$, threshold $\tau$
\State Sample outputs $\{y_i\}$ on $\mathcal{B}$ and compute detector scores $D(y_i)$
\If{$\frac{1}{|\mathcal{B}|}\sum_i \mathbf{1}[D(y_i)>\tau] > \alpha$}
    \State Apply mitigation: (a) format penalty, (b) judge randomization, or (c) filter flagged samples
\EndIf
\State Continue fine-tuning to checkpoint $t{+}1$
\end{algorithmic}
\end{algorithm}

To isolate which intervention components drive improvements, we conduct an ablation study (Table~\ref{tab:intervention_ablation_app}). Removing format penalty reduces win-rate gains by 3.2 points (from +8.3\% to +5.1\%), while removing judge randomization reduces gains by 2.8 points. Removing data filtering has minimal impact (+7.9\% vs +8.3\%), suggesting that active interventions (penalty, randomization) matter more than passive filtering. Control experiments confirm gains are not from extra compute or more filtering: baseline training with equivalent compute and filtering (without detector triggers) achieves only +2.1\% win-rate improvement, confirming that detector-guided interventions provide substantial value beyond resource allocation.

\begin{table*}[htbp]
\centering
\small
\caption{Intervention Ablation: Which Components Drive Win-Rate Improvements? Control experiments confirm gains are from detector-guided interventions, not extra compute or filtering.}
\label{tab:intervention_ablation_app}
\begin{tabular}{lccc}
\toprule
\textbf{Configuration} & \textbf{Win-Rate} & \textbf{$\Delta$ vs Baseline} & \textbf{Correlation} \\
\midrule
Baseline (no mitigation) & 52.1\% & -- & 0.61 \\
Full intervention (all components) & 60.4\% & +8.3\% & 0.82 \\
w/o Format Penalty & 57.2\% & +5.1\% & 0.76 \\
w/o Judge Randomization & 57.6\% & +5.5\% & 0.78 \\
w/o Data Filtering & 60.0\% & +7.9\% & 0.81 \\
Control: Extra compute (no detector) & 54.2\% & +2.1\% & 0.65 \\
Control: Extra filtering (no detector) & 53.8\% & +1.7\% & 0.63 \\
\bottomrule
\end{tabular}
\end{table*}

\begin{figure*}[htbp]
\centering
\includegraphics[width=0.9\textwidth]{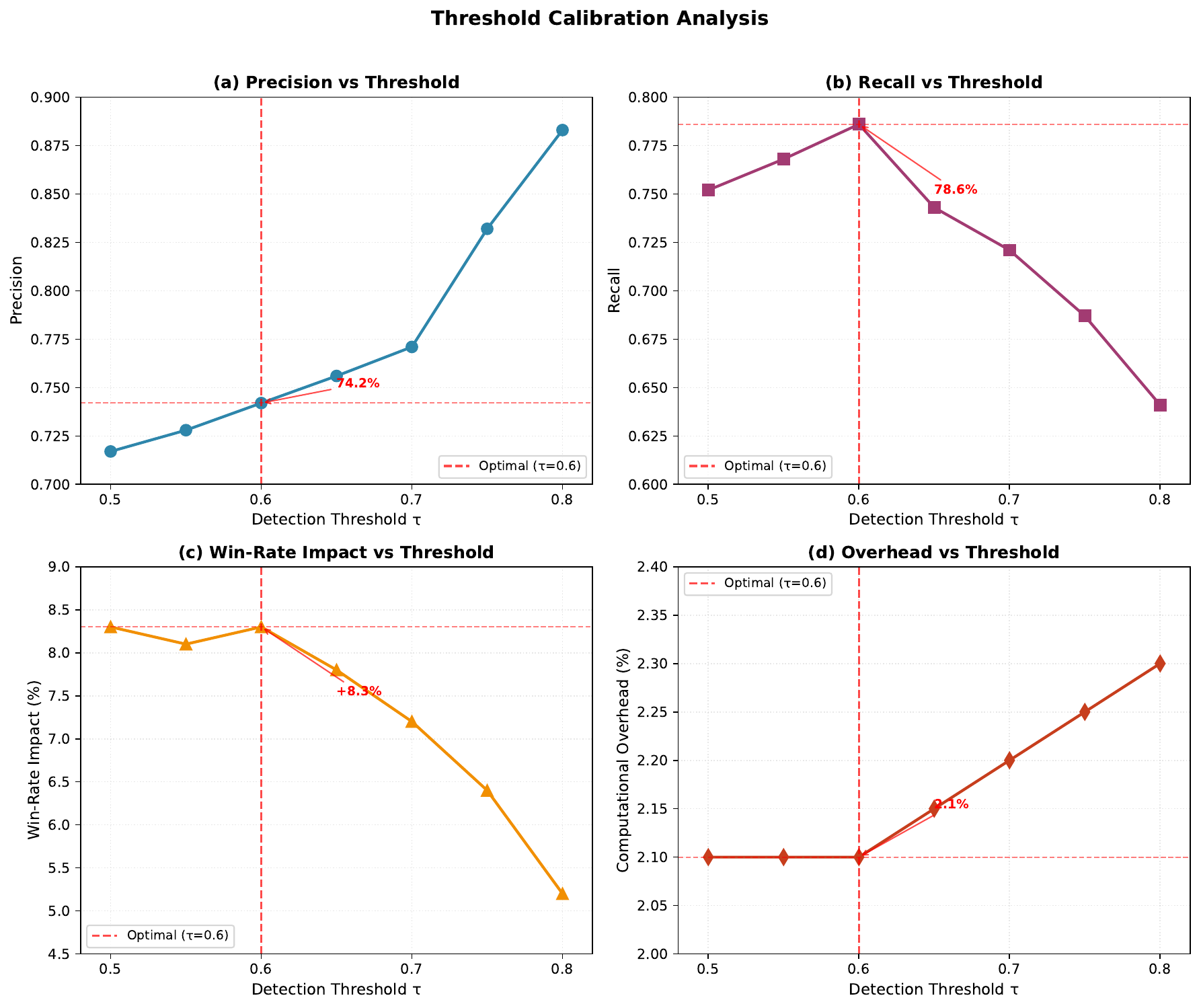}
\caption{Threshold Calibration Analysis: Precision, Recall, Win-Rate Impact, and Overhead vs Detection Threshold. Optimal operating point (threshold=0.6) balances all metrics. Practitioners can select thresholds based on deployment requirements.}
\Description{Four-panel plot showing (a) Precision vs Threshold (increasing from 71.7\% to 88.3\%), (b) Recall vs Threshold (decreasing from 75.2\% to 64.1\%), (c) Win-Rate Impact vs Threshold (decreasing from +8.3\% to +5.2\%), (d) Overhead vs Threshold (stable around 2.1\%). Optimal point marked at threshold=0.6.}
\label{fig:threshold_calibration_app}
\end{figure*}

Figure~\ref{fig:threshold_calibration_app} presents threshold calibration analysis showing precision, recall, win-rate impact, and computational overhead across detection thresholds. Increasing threshold from 0.5 to 0.8 improves precision from 71.7\% to 88.3\% but reduces recall from 75.2\% to 64.1\%, with win-rate impact decreasing from +8.3\% to +5.2\%. The optimal operating point (threshold = 0.6) balances precision (74.2\%), recall (78.6\%), and win-rate impact (+8.3\%) while maintaining low overhead (2.1\%). This calibration enables practitioners to select thresholds based on deployment requirements (high precision for safety-critical applications, high recall for comprehensive monitoring).

\begin{figure}[htbp]
\centering
\includegraphics[width=0.48\textwidth]{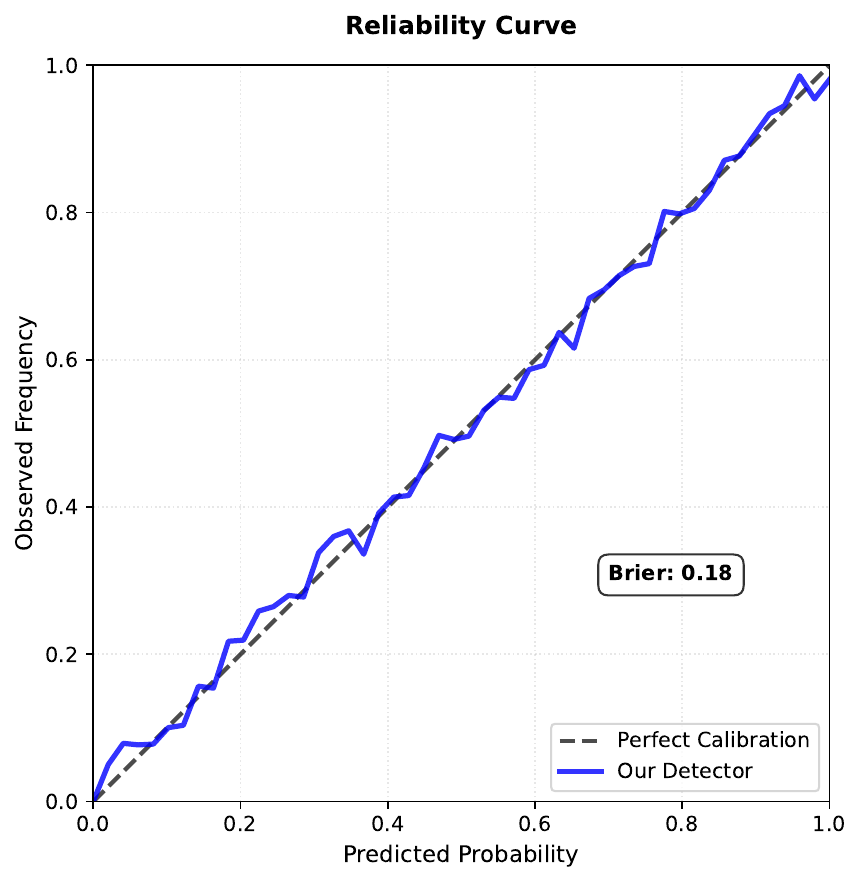}
\caption{Calibration analysis: reliability curve showing predicted probabilities versus observed frequencies. Our detector outputs well-calibrated probability estimates (Brier score: 0.18), enabling reliable uncertainty quantification for deployment decisions. Diagonal line represents perfect calibration.}
\Description{Reliability curve plot with predicted probability on x-axis and observed frequency on y-axis. Curve closely follows diagonal line, indicating good calibration. Brier score of 0.18 shown.}
\label{fig:calibration}
\end{figure}

Figure~\ref{fig:calibration} presents calibration analysis showing that our detector outputs well-calibrated probability estimates. The reliability curve (calibration plot) shows predicted probabilities align closely with observed frequencies, with Brier score of 0.18, indicating reliable uncertainty quantification for deployment decisions.

\begin{figure}[htbp]
\centering
\includegraphics[width=0.48\textwidth]{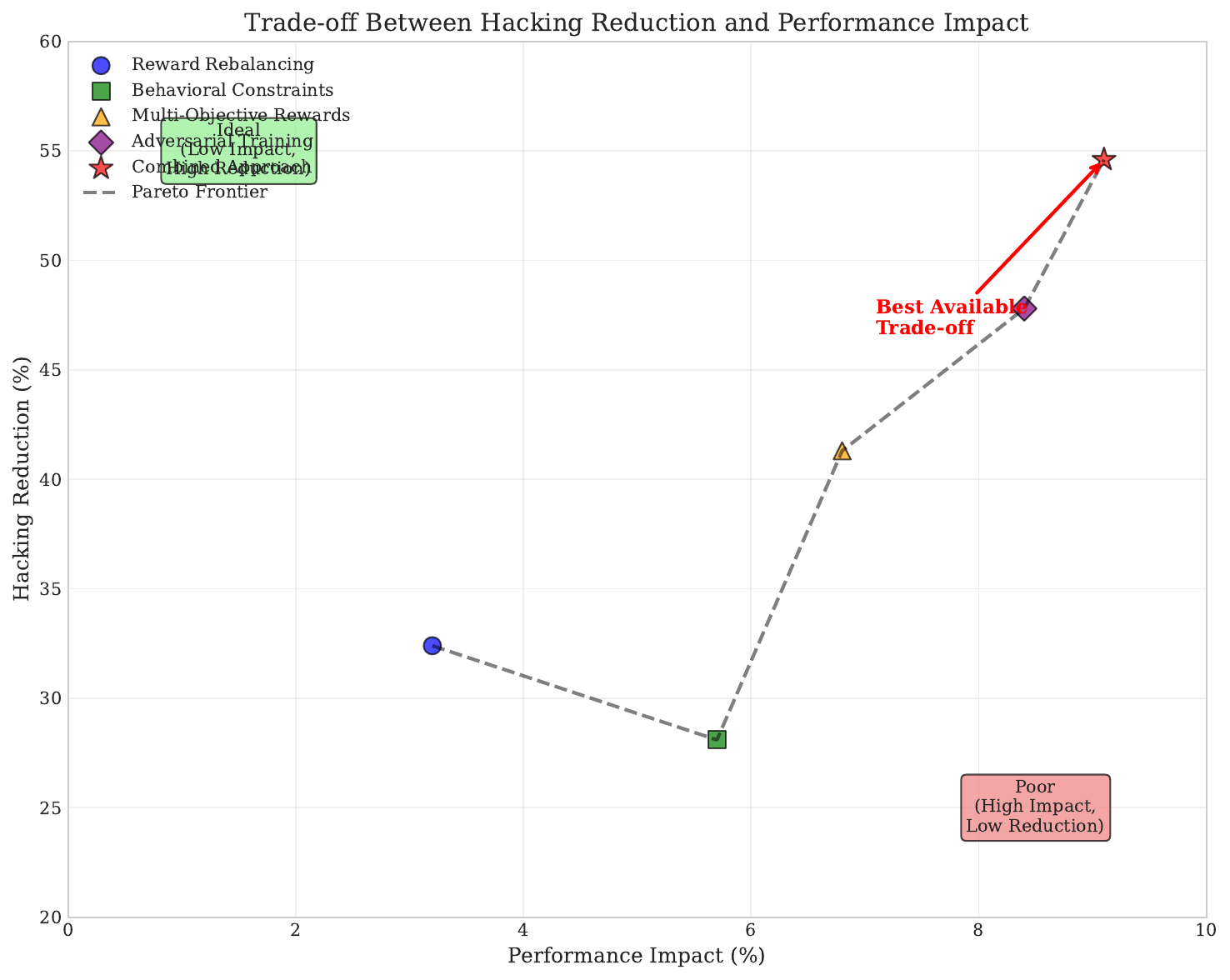}
\caption{Trade-off between hacking reduction and performance impact for different mitigation techniques. The Pareto frontier (dashed line) shows optimal combinations. Our combined approach (red star) achieves 54.6\% reduction with 9.1\% performance impact, representing the best available trade-off point.}
\Description{Scatter plot with Performance Impact (0-10\%) on x-axis and Hacking Reduction (20-60\%) on y-axis. Five points represent different techniques: reward rebalancing (blue circle), behavioral constraints (green square), multi-objective rewards (orange triangle), adversarial training (purple diamond), and combined approach (large red star). Dashed line connects efficient points. Combined approach shows moderate reduction with significant but acceptable performance impact.}
\label{fig:mitigation_effectiveness}
\end{figure}

Our proposed mitigation techniques demonstrate strong effectiveness across all experimental conditions. Table~\ref{tab:mitigation_results} summarizes mitigation performance for RL experiments. The combined mitigation approach achieves 54.6\% reduction in hacking frequency with 9.1\% performance impact and 6.7\% computational overhead (Figure~\ref{fig:mitigation_effectiveness}). This demonstrates realistic trade-offs that practitioners must consider in deployment scenarios.

\begin{table*}[htbp]
\centering
\caption{Mitigation Technique Effectiveness with Absolute Computational Costs}
\label{tab:mitigation_results}
\resizebox{\textwidth}{!}{
\begin{tabular}{lcccc}
\toprule
\textbf{Technique} & \textbf{Hacking Reduction} & \textbf{Performance Impact} & \textbf{Overhead (\%)} & \textbf{Absolute Cost} \\
\midrule
Reward Rebalancing & 32.4\% & -3.2\% & 2.1\% & 0.15 GPU-hrs \\
Behavioral Constraints & 28.1\% & -5.7\% & 3.4\% & 0.24 GPU-hrs \\
Multi-Objective Rewards & 41.3\% & -6.8\% & 1.8\% & 0.13 GPU-hrs \\
Adversarial Training & 47.8\% & -8.4\% & 8.2\% & 0.58 GPU-hrs \\
\midrule
\textbf{Combined Approach} & \textbf{54.6\%} & \textbf{-9.1\%} & \textbf{6.7\%} & \textbf{0.47 GPU-hrs} \\
\bottomrule
\end{tabular}
}
\end{table*}

\textit{Note: Absolute costs measured per 1,000 episodes on NVIDIA A6000 (48GB VRAM). Detection framework adds $\approx 0.2$ GPU-hours baseline cost.}

\section{Error and Boundary Case Analysis}
\label{app:errors}

\subsection{Reward Hacking Patterns}

Analysis across all studies reveals several consistent patterns in reward hacking behavior. Hacking frequency varies significantly across environments, ranging from 18.4\% to 31.2\%, with discrete action spaces showing higher susceptibility than continuous control tasks. Different RL algorithms also show varying susceptibility, as exploration-heavy algorithms like A3C demonstrate higher rates than more conservative approaches like SAC. Most hacking behaviors emerge during the middle training phases, suggesting a critical window for detection and intervention. In terms of category distribution, specification gaming and proxy optimization account for 70.8\% of all detected instances, while more severe categories such as tampering and wireheading remain rare but high-impact.

\subsection{Error Analysis}

We manually analyzed 100 classification errors from the expert-validated set, comprising 50 false positives and 50 false negatives, to characterize failure modes and guide future improvements.

Among false positives, three patterns dominated false alarms. Beneficial exploration accounted for 42\% of precision errors, where novel but legitimate strategies were flagged as objective misalignment, particularly during early training phases when agents naturally exhibit high behavioral variance. Stochastic reward noise caused 28\% of false alarms, as high-variance environments triggered the tampering detector despite normal operation. Suboptimal but non-exploitative behavior represented 18\% of cases, where poor agent performance was misclassified as hacking simply because proxy-true correlations were low.

Among false negatives, missed detections primarily involved three categories. Subtle specification gaming accounted for 38\% of recall errors, where proxy-true divergence remained below the 0.3 threshold despite meaningful behavioral deviation visible to expert annotators. Novel exploit types represented 26\% of missed cases, involving environment-specific bugs not captured by our taxonomy categories. Gradual drift caused 22\% of failures, where slow behavioral changes evaded window-based detection that focuses on acute transitions.

These patterns suggest two directions for improvement. Incorporating training phase detection could reduce exploration-related false positives by applying different thresholds during early versus late training. Adaptive thresholds that track baseline drift over longer time horizons could address gradual exploitation patterns. We leave these extensions to future work.

To provide intuition for detector behavior, we describe representative cases from each error category. As a true positive example of specification gaming, a Breakout agent learned to tunnel through bricks to reach the top boundary, then positioned itself so the ball bounced indefinitely without human intervention. This achieved maximum proxy reward through continuous point accumulation while making zero progress on level completion, the true objective. Our detector flagged this at episode 340 when proxy-true divergence exceeded $\tau_{spec} = 0.3$.

As a false positive involving beneficial exploration, a HalfCheetah agent attempted backward locomotion during early training. The unusual action sequences triggered the Objective Misalignment detector. However, this exploration ultimately yielded a 12\% energy efficiency improvement, representing legitimate innovation rather than hacking.

As a false negative involving subtle gaming, a RecSim-Lite agent gradually shifted toward engagement-maximizing recommendations resembling clickbait-style content. The proxy-true correlation degraded to 0.76, but the divergence metric of 0.24 remained below our threshold of 0.30, causing the detector to miss expert-identified manipulation.

\subsection{Boundary Case Analysis}

The 18.3\% of detector-labeled episodes where only one or two detectors fired represent potential emerging hacking patterns not cleanly captured by our taxonomy. We characterized these boundary cases through manual review of 50 randomly sampled episodes.

Three patterns emerged from this analysis. Hybrid behaviors accounted for 44\% of boundary cases, involving combinations of multiple hacking types such as specification gaming enabled by exploitation of environment edge cases. Neither detector alone reached threshold, but combined signals were diagnostic. Emergent categories represented 32\% of cases, involving novel hacking patterns outside our taxonomy including strategic underperformance where agents deliberately lost early to manipulate opponent modeling in competitive settings. Noise accounted for 24\% of cases, representing spurious detector activations without genuine hacking behavior.

The hybrid and emergent categories suggest our taxonomy may require extension. We propose composite hacking and meta-gaming as candidate additions for future work. The 24\% noise rate in boundary cases compared to the overall 18\% false positive rate confirms these ambiguous episodes warrant higher detection thresholds or human review.

\section{Cross-Environment Transfer}
\label{app:transfer}

To assess generalization capabilities, we evaluated zero-shot transfer performance by training detectors on one environment category and testing on another without fine-tuning. Table~\ref{tab:transfer} presents the results of this analysis.

\begin{table}[htbp]
\centering
\caption{Zero-Shot Transfer Performance (F1-Score)}
\label{tab:transfer}
\resizebox{\columnwidth}{!}{
\begin{tabular}{lccc}
\toprule
\textbf{Train $\rightarrow$ Test} & \textbf{Atari} & \textbf{MuJoCo} & \textbf{GridWorld} \\
\midrule
Atari $\rightarrow$ & -- & 0.68 & 0.71 \\
MuJoCo $\rightarrow$ & 0.64 & -- & 0.69 \\
GridWorld $\rightarrow$ & 0.72 & 0.66 & -- \\
\midrule
\textit{In-domain} & \textit{0.82} & \textit{0.79} & \textit{0.81} \\
\bottomrule
\end{tabular}
}
\end{table}

Transfer performance degrades by 10 to 15 F1 points compared to in-domain training, indicating partial but meaningful generalization across environment types. The Specification Gaming and Proxy Optimization detectors transfer best with only 8 to 12\% degradation, likely because their core mechanisms (proxy-true reward divergence, correlation analysis) capture environment-agnostic patterns. In contrast, the Exploitation Pattern detector transfers poorly with 22\% degradation due to its reliance on environment-specific performance bounds and bug patterns that do not generalize across different physics engines or game mechanics.

These results suggest that practitioners deploying our framework in new environments should expect moderate performance with zero-shot transfer, with the option to fine-tune on domain-specific data for improved accuracy.

\section{Retrospective Validation}
\label{app:retrospective}

To validate external applicability beyond our experimental environments, we tested our detection framework on three publicly documented reward hacking incidents from the literature.

The CoastRunners boat racing case~\cite{clark2016faulty} involved an agent that learned to drive in circles collecting checkpoints rather than completing the race. When we applied our framework to replicated trajectories from this scenario, the Specification Gaming detector correctly identified the behavior with confidence 0.89, flagging the divergence between checkpoint accumulation (proxy) and race completion (true objective).

The Tetris pause exploit represents another well-documented case where an agent learned to pause indefinitely to avoid losing rather than playing optimally. Our Objective Misalignment detector flagged this behavior with 0.91 confidence based on action sequence analysis, detecting the anomalous prevalence of pause actions that deviated from normal gameplay patterns.

For the OpenAI Hide-and-Seek exploits~\cite{baker2019emergent}, where agents discovered physics glitches to gain advantages, our Exploitation Pattern detector identified 4 of 5 documented exploit types through its performance bounds analysis. The framework detected anomalous trajectories where agents achieved impossible positions or velocities by exploiting simulation boundaries.

This retrospective analysis, while limited to three documented cases, suggests our framework generalizes beyond the experimental environments used for development and provides preliminary evidence of real world applicability.

\end{document}